%% file: main_w_appendix.tex
\documentclass[10pt,twocolumn,letterpaper]{article}

\usepackage{cvpr}              

\input{preamble}

\definecolor{cvprblue}{rgb}{0.21,0.49,0.74}
\usepackage[pagebackref,breaklinks,colorlinks,allcolors=cvprblue]{hyperref}
\usepackage[accsupp]{axessibility}
\usepackage{url}
\usepackage[utf8]{inputenc} 
\usepackage[T1]{fontenc}    
\usepackage{hyperref}       
\usepackage{url}            
\usepackage{booktabs}       
\usepackage{amsfonts}       
\usepackage{nicefrac}       
\usepackage{microtype}      
\usepackage{xcolor}         
\usepackage{graphicx}
\usepackage{epstopdf}
\usepackage{subfloat}
\usepackage{multirow}
\usepackage{amsmath}
\usepackage{algorithm}
\usepackage{algorithmic}
\usepackage{listings}
\usepackage{enumitem}
\usepackage{bbm}
\usepackage{wrapfig}   
\usepackage{caption}   
\usepackage[normalem]{ulem}

\usepackage{colortbl}
\usepackage{listings}

\def\paperID{14597} 
\def\confName{CVPR}
\def\confYear{2025}

\title{From Head to Tail: Towards Balanced Representation in Large Vision-Language Models through Adaptive Data Calibration}

\renewcommand{\thefootnote}{\dag}
\makeatletter
\renewcommand\@fnsymbol[1]{%
  \ifcase#1\or \dag\else \@arabic{#1}\fi}
\makeatother

\author{
Mingyang Song$^{1,2}$, Xiaoye Qu$^2$\footnotemark[1], Jiawei Zhou$^3$\thanks{Corresponding authors}, Yu Cheng$^{4}$\footnotemark[1] \\
$^{1}$Fudan University, 
$^{2}$Shanghai Artificial Intelligence Laboratory\\
$^{3}$Stony Brook University, $^{4}$The Chinese University of Hong Kong\\
\texttt{mysong23@m.fudan.edu.cn}; 
\texttt{quxiaoye@pjlab.org.cn}; \\
\texttt{jiawei.zhou.1@stonybrook.edu};
\texttt{chengyu@cse.cuhk.edu.hk};\\
Project Page: \href{https://vlmlt.github.io/}{\texttt{%
    \textcolor[rgb]{0.7,0.1,0.1}{h}%
    \textcolor[rgb]{0.8,0.4,0.1}{t}%
    \textcolor[rgb]{0.7,0.7,0.0}{t}%
    \textcolor[rgb]{0.1,0.5,0.1}{p}%
    \textcolor[rgb]{0.1,0.5,0.5}{s}%
    \textcolor[rgb]{0.1,0.1,0.7}{:}%
    \textcolor[rgb]{0.5,0.1,0.5}{/}%
    \textcolor[rgb]{0.5,0.1,0.5}{/}%
    \textcolor[rgb]{0.7,0.1,0.1}{v}%
    \textcolor[rgb]{0.8,0.4,0.1}{l}%
    \textcolor[rgb]{0.7,0.7,0.0}{m}%
    \textcolor[rgb]{0.1,0.5,0.1}{l}%
    \textcolor[rgb]{0.1,0.5,0.5}{t}%
    \textcolor[rgb]{0.1,0.1,0.7}{.}%
    \textcolor[rgb]{0.5,0.1,0.5}{g}%
    \textcolor[rgb]{0.7,0.1,0.1}{i}%
    \textcolor[rgb]{0.8,0.4,0.1}{t}%
    \textcolor[rgb]{0.1,0.5,0.5}{h}%
    \textcolor[rgb]{0.1,0.1,0.7}{u}%
    \textcolor[rgb]{0.5,0.1,0.5}{b}%
    \textcolor[rgb]{0.7,0.1,0.1}{.}%
    \textcolor[rgb]{0.8,0.4,0.1}{i}%
    \textcolor[rgb]{0.7,0.7,0.0}{o}%
}}
}

\begin{document}
\maketitle

\input{Styles/content/0.abstract}

\input{Styles/content/1.intro}

\input{Styles/content/2.analysis}

\input{Styles/content/3.approach}

\input{Styles/content/4.experiment}

\input{Styles/content/5.ablation}

\input{Styles/content/6.conclusion}

{
    \small
    \bibliographystyle{ieeenat_fullname}
    \bibliography{main_w_appendix}
}

\clearpage
\input{Styles/content/8.5.new_appendix}

\end{document}

%% file: preamble.tex
\newcommand{\red}[1]{{\color{red}#1}}


%% file: Styles/content/0.abstract.tex
\begin{abstract}

Large Vision-Language Models (LVLMs) have achieved significant progress in combining visual comprehension with language generation.
Despite this success, the training data of LVLMs still suffers from \textit{Long-Tail (LT)} problems, where the data distribution is highly imbalanced.
Previous works have mainly focused on traditional VLM architectures, i.e., CLIP or ViT, and specific tasks such as recognition and classification. Nevertheless, the exploration of LVLM~(e.g. LLaVA) and more general tasks~(e.g. Visual Question Answering and Visual Reasoning) remains under-explored.
In this paper, we first conduct an in-depth analysis of the LT issues in LVLMs and identify two core causes: the overrepresentation of head concepts and the underrepresentation of tail concepts.
Based on the above observation, we propose an \textbf{A}daptive \textbf{D}ata \textbf{R}efinement Framework (\textbf{ADR}), which consists of two stages: \textbf{D}ata \textbf{R}ebalancing (\textbf{DR}) and \textbf{D}ata \textbf{S}ynthesis (\textbf{DS}).
In the DR stage, we adaptively rebalance the redundant data based on entity distributions, while in the DS stage, we leverage Denoising Diffusion Probabilistic Models (DDPMs) and scarce images to supplement underrepresented portions.
Through comprehensive evaluations across eleven benchmarks, our proposed ADR effectively mitigates the long-tail problem in the training data, improving the average performance of LLaVA 1.5 relatively by 4.36\%, without increasing the training data volume. 
\end{abstract}

%% file: Styles/content/1.intro.tex
\begin{figure}[t!]
\centering    
\subfloat[Performance over LLaVA 1.5.] {
\includegraphics[width=0.49\linewidth]{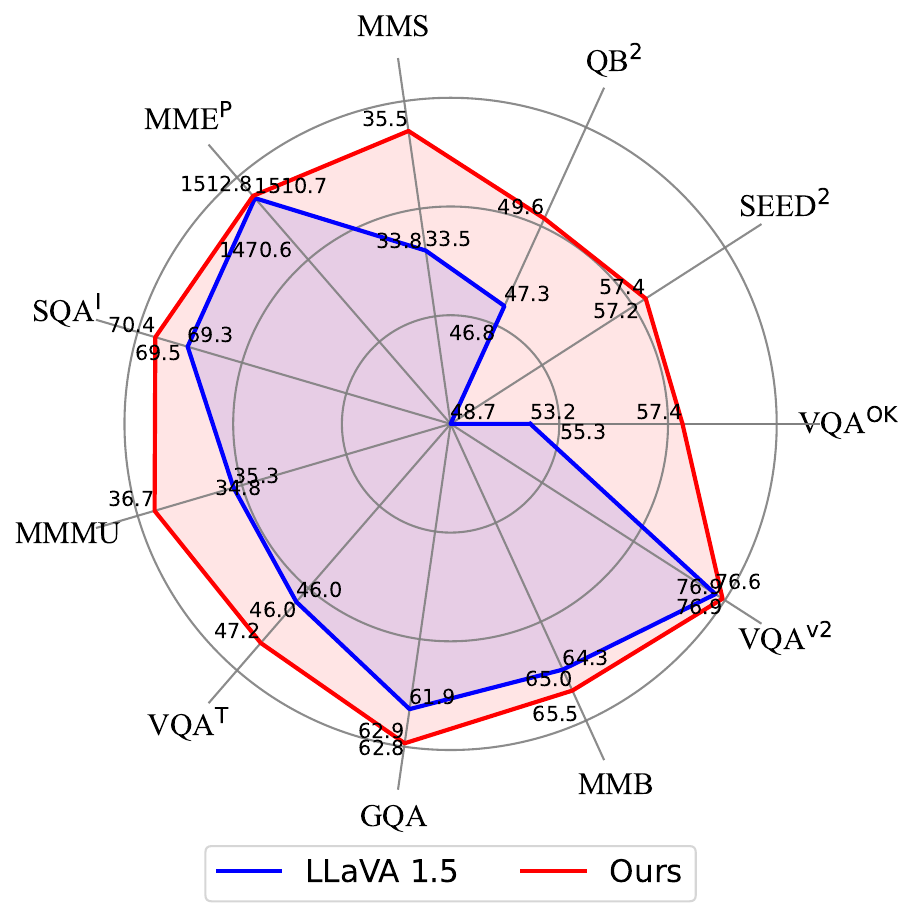}  
}     
\subfloat[Tail 30\% Acc. on VQAV2.] {  
\includegraphics[width=0.45\linewidth]{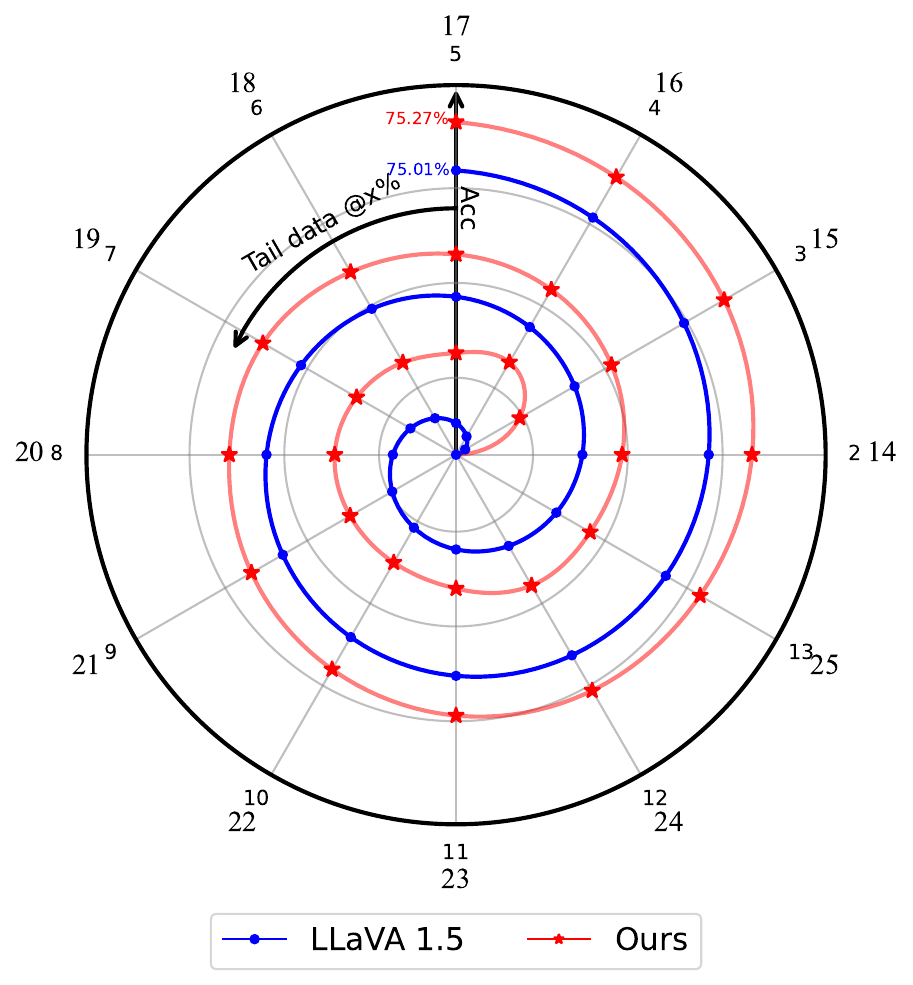} 
}
\caption{
Performance before and after addressing the LT problem. Our method surpasses the baseline over all benchmarks and also effectively improves the performance of tail \textbf{30\%} concepts.
}     
\label{fig:head_picture}     
\vspace{-10pt}
\end{figure}

\begin{figure*}[!t]
    \centering
    \includegraphics[width=0.95\linewidth]{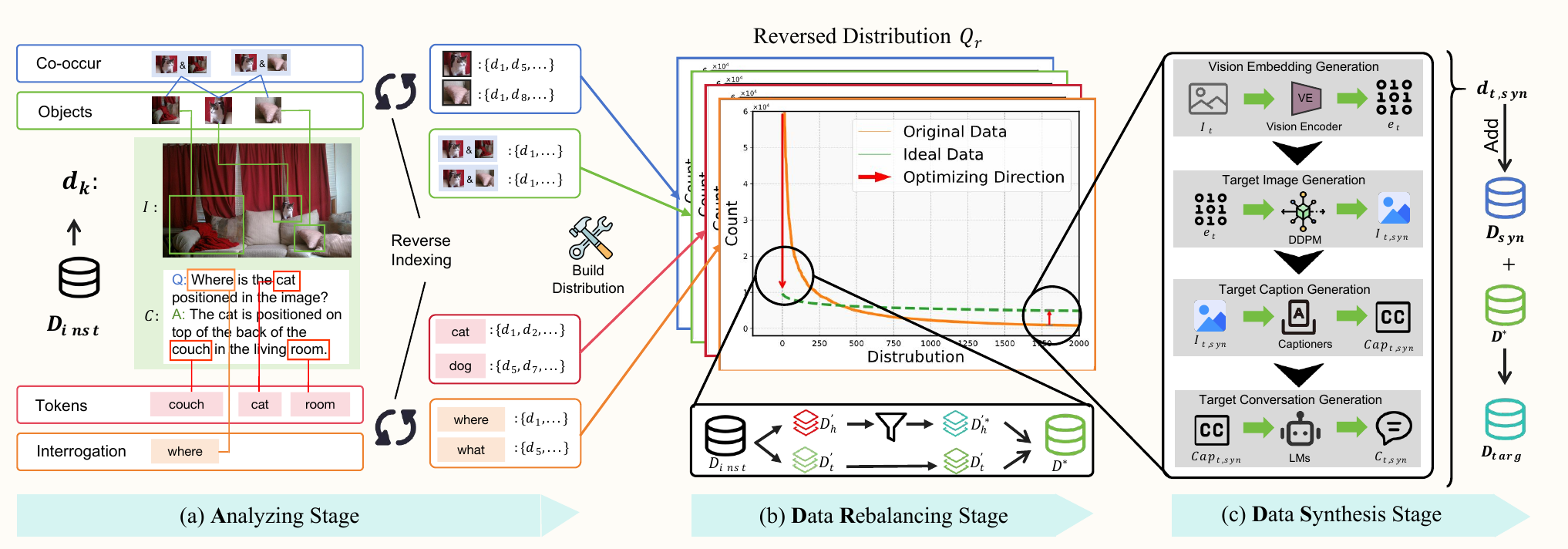}
    \caption{
    The overview of our Adaptive Data Refinement Framework (ADR). 
    (a) In the {Analyzing} Stage, we first extract tokens, objects, co-occurrences, and interrogations from the training instances, then construct corresponding distribution using a reverse-indexed mapping. 
    (b) In the Data Rebalancing stage, we analyze the optimizing direction and adaptively rebalance the redundant data based on the entity distribution identified in the Analyzing stage. 
    (c) Finally, in the Data Synthesis stage, we utilize DDPM and the latent representations of scarce image instances to synthesize the underrepresented data.
    }
    \label{fig:main_figure}
    \vspace{-10pt}
\end{figure*}

\section{Introduction}

Large Vision-Language Models (LVLMs) have become pivotal at the intersection of computer vision and natural language processing, facilitating a wide range of applications. Recent advancements in LVLMs \citep{Qwen-VL,chen2023shikra,dai2024instructblip,zhang2023internlmxcomposer,dong2024internlmxcomposer2,chen2023internvl,liu2023improvedllava,liu2024visual,zhu2023minigpt,ye2023mplugowl,abdin2024phi3,qu2024alleviating,qu2024mitigating,qu2024look,liu2024survey} have significantly advanced general-purpose foundation models, elevating them to unprecedented levels. However, the training data of LVLMs are suffering from the problem of \textit{Long-Tail (LT) } \citep{parashar2024neglected}, which refers to the fact that the training datasets present highly imbalanced distributions, and they have a large number of tail classes. 

As a widely existing phenomenon in real-world distribution, some recent research \citep{zhang2024ltreview,zhang2023ltsurvey2,fu2022ltsurvey3,yang2022ltsurvey4} found that balancing the LT data can bring positive effects. 
To achieve it, several works \citep{shao2023investigating_cliplt1,wang2022debiased_cliplt2,zhu2024generalized_cliplt3,liu2024lessismore,lee2024concept} attempt to filter the redundant data or re-design the network structure. However, these studies primarily focus on traditional models (e.g., CLIP \citep{radford2021learning_CLIP}) or tasks (e.g., image classification). In contrast, the LT problem in LVLMs presents unique challenges due to their cross-modal nature, the involvement of multiple aspects, and the distinctive co-occurrence phenomenon, which still remains under-explored.

To address the above issue, in this paper, we delve into the distribution of the training data, analyzing it from four perspectives based on both language and visual features, and introduce an \textbf{A}daptive \textbf{D}ata \textbf{R}efinement Framework (\textbf{ADR}). 
Our framework can be easily integrated into training data of any open-source LVLMs such as LLaVA \citep{liu2023improvedllava}, ShareGPT-4V \citep{chen2023sharegpt4v}, ALLaVA \citep{chen2024allava}, or Mini-GPT4V \citep{zhu2023minigpt}.
Specifically, ADR addresses the long-tail (LT) problem by both filtering redundant data and synthesis scarce data through three key stages: \textbf{A}nalyzing Stage, \textbf{D}ata \textbf{R}ebalancing Stage (DR) and  \textbf{D}ata \textbf{S}ynthesis Stage (DS). 
Firstly, the Analyzing stage assesses the severity of LT problem and builds key entity distributions from four different perspectives, including tokens, objects,
co-occurrences, and interrogations.
Then, the DR stage rebalances the overrepresented head portion of the data by filtering out low-quality or redundant instances based on entity distributions, thereby mitigating overfitting to redundant head data. Finally, the DS stage leverages the latent representations of scarce images to adaptively synthesize the underrepresented tail data, significantly improving the performance on tail-end concepts of the model.

To comprehensively evaluate our framework, we adopt diverse general-purpose benchmarks, hallucination metrics, and GPT-4 evaluations to validate the effectiveness.  
The results demonstrate that our proposed ADR framework significantly improves performance over baseline data and methods. As shown in Figure 1(a), across all 11 benchmarks, ADR consistently achieves better results, improving the average performance of LLaVA 1.5 relatively by 4.36\%, without increasing the training data volume or introducing additional training stages. Furthermore, the performance on the tail data also observes a significant improvement as depicted in Figure 1(b).
To sum up, our contributions are threefold:

\setlength{\itemsep}{0pt}
\begin{itemize}[leftmargin=*]
\setlength{\itemsep}{0pt}

   \item  We present the first in-depth analysis of the unique LT challenges in LVLMs from four key aspects across both vision and language modalities, including tokens, objects,
   co-occurrences, and interrogations. Our analysis reveals the significant LT problem within LVLMs' training data.

   \item We introduce a comprehensive Adaptive Data Refinement (ADR) framework to effectively mitigate the LT problem without increasing data volume or requiring additional training. ADR is both model-agnostic and data-agnostic, making it easily transferable to any open-source dataset.

    \item Comprehensive evaluation over general-purpose, hallucination, and GPT assessments proves the superiority of our framework. Moreover, we present the potential of our framework to serve as a general toolkit as it can be easily transferred to any open-source data.

\end{itemize}

\begin{figure*}[t!]
\vspace{-10pt}
\centering    
\subfloat[MME-Tok]{
\includegraphics[width=0.24\linewidth]{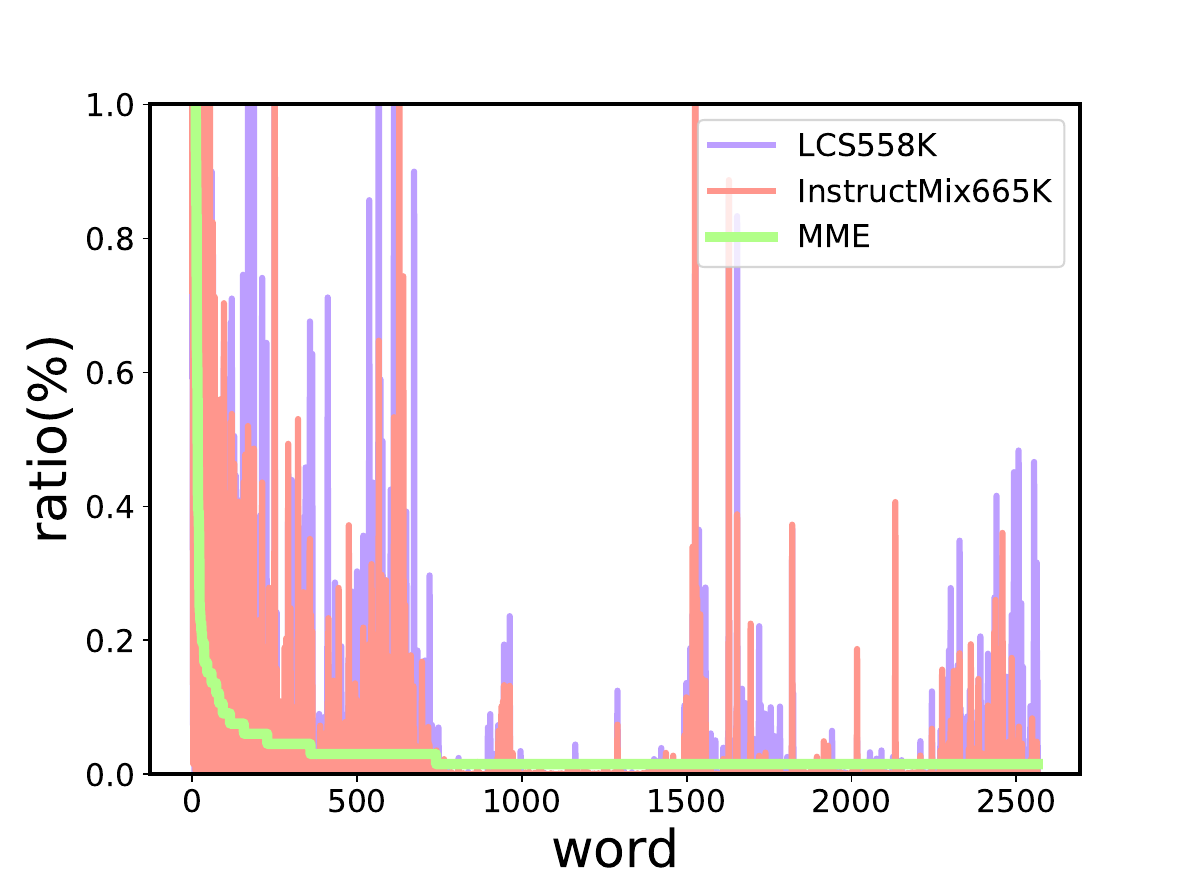}  
}     
\subfloat[LLaVA-Tok] { 
\includegraphics[width=0.24\linewidth]{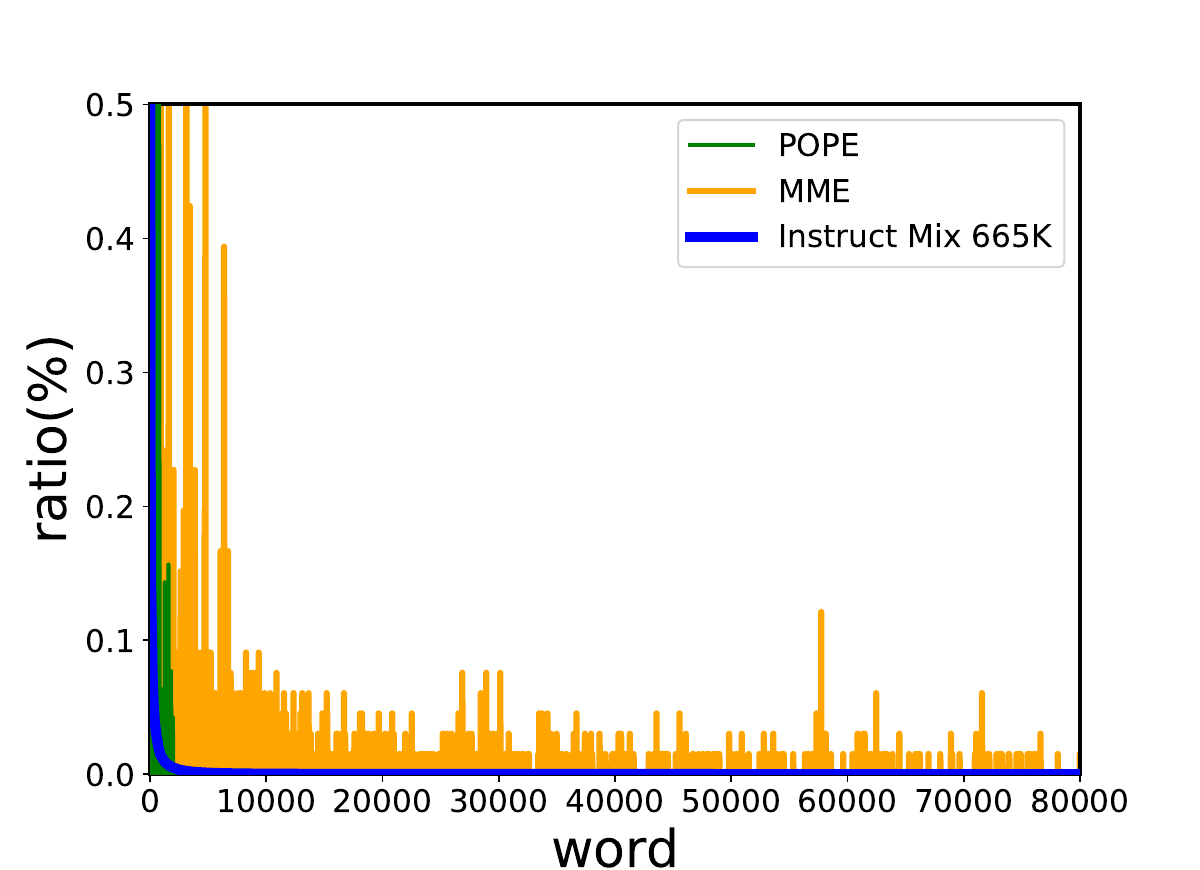}     
}
\subfloat[MME-Obj] {  
\includegraphics[width=0.24\linewidth]{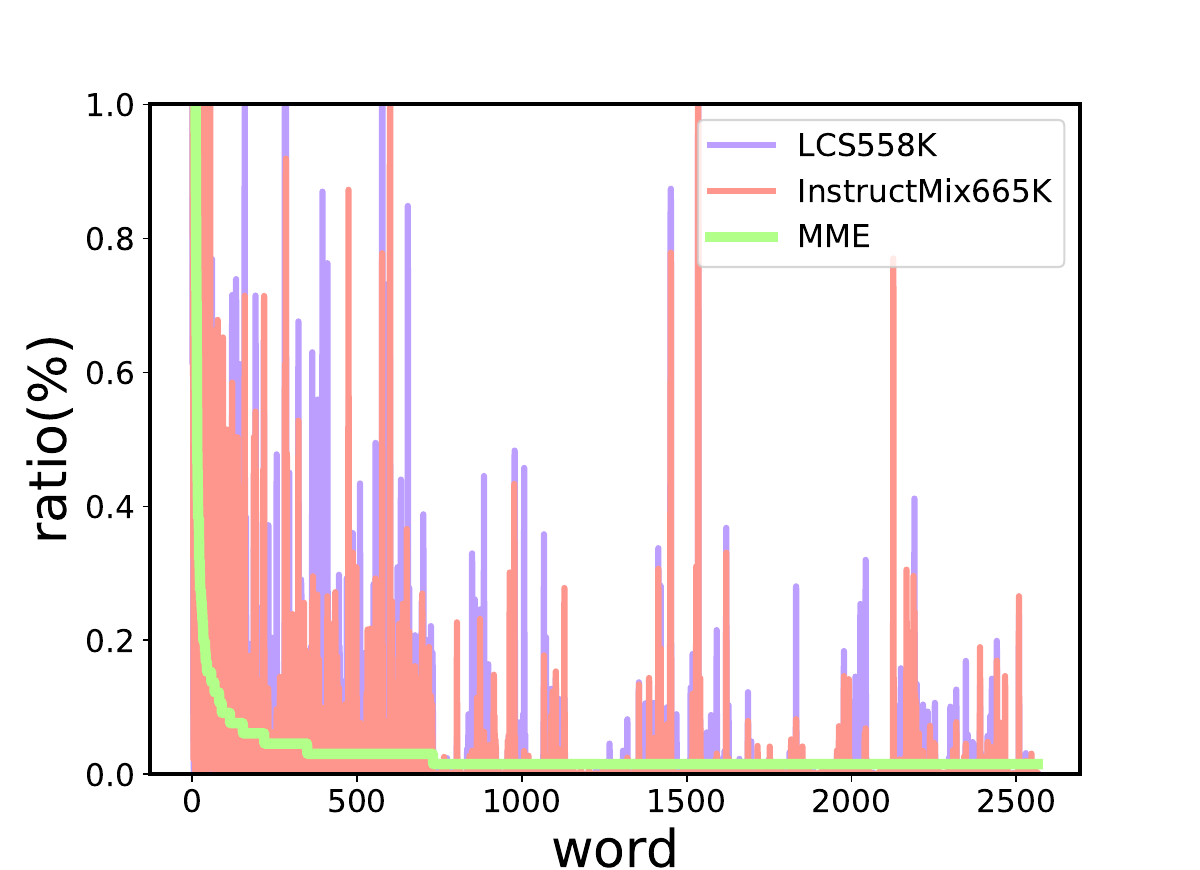}  
}     
\subfloat[LLaVA-Obj] { 
\includegraphics[width=0.24\linewidth]{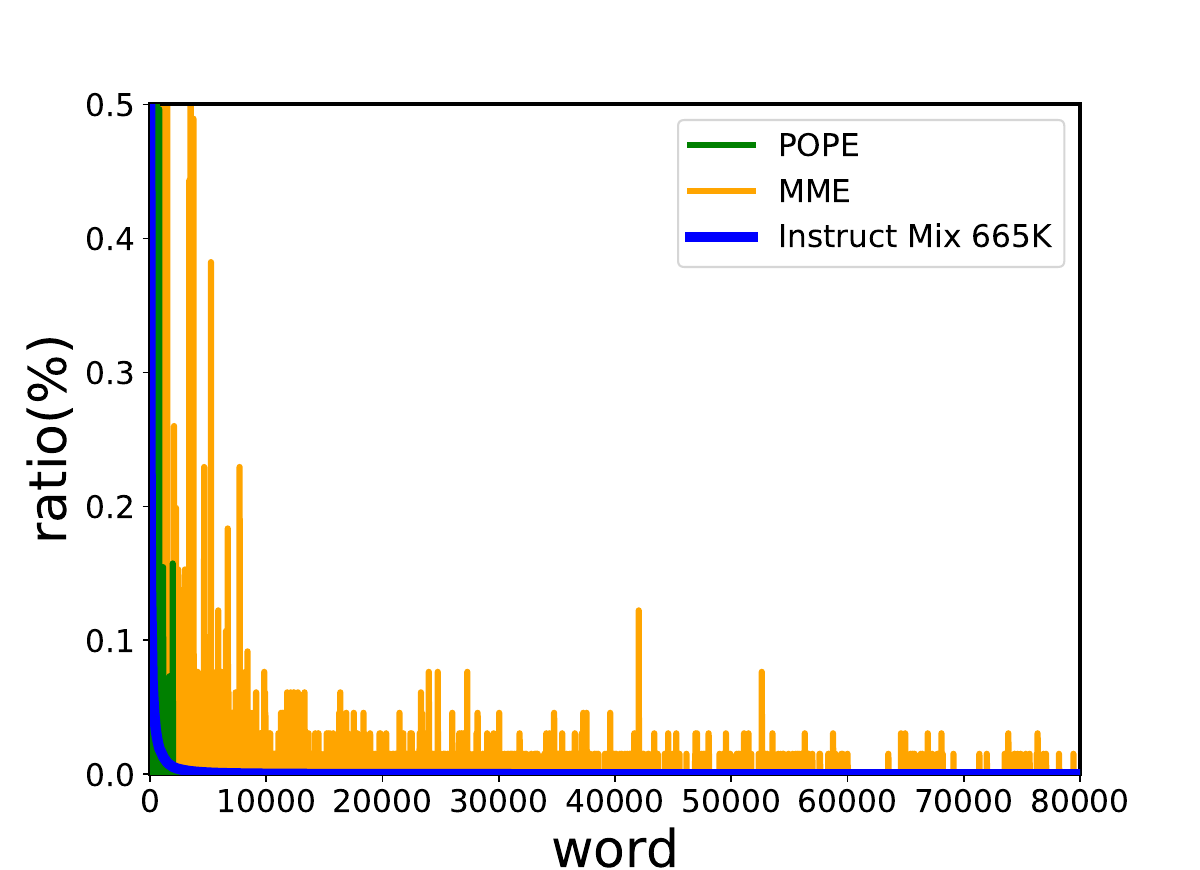}    
}
 
\caption{ 
Long-tail distribution in instruction-tuning and benchmark datasets:
(a) Token-level distribution in MME \citep{fu2023MME}.
(b) Token-level distribution in InstructMix665K \citep{liu2024visual}.
(c) Object-level distribution in MME \cite{fu2023MME}.
(d) Object-level distribution in InstructMix665K \citep{liu2024visual}.
}
\label{fig:long_tail_distribution_analysis} 
\vspace{-10pt}
\end{figure*}

\section{Related Work}
\subsection{Large Vision-Language Models}

Recently, Large Vision-Language Models (LVLMs) have attracted significant attention. Besides the powerful business models such as GPT-4V, GPT-4o, Gemini, and Claude \citep{openai2023gpt4v,openai2024gpt4o,team2023gemini,anthropic2024claude3}, many open-source LVLMs emerge \citep{liu2023improvedllava,Qwen-VL,zhu2023minigpt,dai2024instructblip}. 
With the aid of strong large language models such as LLaMA \citep{touvron2023llama} or Vicuna \citep{vicuna2023}, and a powerful vision encoder such as CLIP \citep{radford2021learning_CLIP}, they manage to align visual comprehension with remarkable language generation capabilities. 
However, all of the aforementioned LVLMs still face significant LT issues regardless of their structure or training phases. Therefore, this paper focuses on addressing the LT problems to facilitate the practical application of LVLMs.

\subsection{Data Development of LVLMs}

The instruction-tuning data for LVLMs typically includes carefully crafted instructions designed to enhance the general instruction-following capabilities or improve downstream task performance of LVLMs. These instructions are often generated by large language models (LLMs) like GPT-4 \citep{liu2024visual} or LVLMs such as GPT-4V \citep{chen2024allava,yan2024listitembyone,tang2024textsquare,su2024conflictbank, su2024timo}. 
Notably, ShareGPT4V \citep{chen2023sharegpt4v} is initially developed from 100K high-quality captions collected from GPT-4V, which are later expanded to 1.2M using a captioning model.
Additionally, various data augmentation techniques are employed during LVLM development, such as random cropping and flipping for vision encoders \citep{ye2024mplugowl2} and projectors \citep{li2023blip-2,ye2023mplugdocowl}, as well as word- and sentence-level augmentation for instruction tuning \citep{chen2024visualitwithpoliteflamingo}. However, these augmentation methods often overlook the inherent distribution of the training data, leading to an inability to balance the data distribution effectively.

Besides data acquisition and augmentation, there is also significant research on data filtering. 
For instance, \citet{paul2021diet_el2n} introduces two popular importance scores for effective data pruning. 
TIVE \citep{liu2024lessismore} leverages gradient-based importance scores to design a data filtering pipeline.
COINCIDE \citep{lee2024concept} examines the distributional differences between training sets and benchmark data, using a smaller model as a reference to select visual instruction tuning data for more efficient fine-tuning of the target LVLM.

\subsection{Long Tail Analysis of VLMs}
Some recent studies \citep{shao2023investigating_cliplt1,wang2022debiased_cliplt2,zhu2024generalized_cliplt3,zhao2024ltgc,zhang2024learning,zhang2024clip,su2024living} seek to mitigate imbalanced predictions of VLMs by training on additional data from downstream tasks. For example, MetaCLIP \citep{xu2023metaclip} analyzes the long tail problem of CLIP pre-training data and uses sub-string matching and inverted indexing to balance the pre-training dataset. GENSYNTH \citep{smith2023balancegender} generates balanced contrast sets through image editing to mitigate spurious correlations in vision-language models.
REAL \citep{parashar2024neglected} analyzes the long tail problem of popular image recognition datasets and designs a tail concept replacement method during the inference stage, significantly improving the recognition accuracy of VLMs. 
However, the long-tail problem within generative LVLMs is still under-explored.

%% file: Styles/content/2.analysis.tex
\begin{table}
\centering
\caption{Relative data volume of tail data after reverse indexing. ``Tok'', ``Obj'', ``Co'', and ``Int'' represent Token, Object, Co-occurrence, and Interrogation, respectively. \textbf{\%E} denotes the percentage of tail entities, while \textbf{\%DI} indicates the percentage of tail data instances.}
\label{tab:reverse-indexing}
\setlength{\extrarowheight}{-2pt}
\resizebox{0.7\columnwidth}{!}{
\begin{tabular}{ccccc}
\toprule
\textbf{Data} & \textbf{Level} & \textbf{thres} & \textbf{\% E} & \textbf{\% DI}\\
\midrule

\multirow{4}{*}{\textbf{LLaVA} \cite{liu2023improvedllava}} &    Tok        & 120   & 98.7  & 10.0  \\ 
                                      & Obj       & 304   & 98.0  & 10.0  \\
                                      & Co    & 24    & 92.7  & 25.0  \\
                                      & Int & 4895  & 99.6  & 10.0  \\
                                \hline
                                      \cellcolor{gray!10}Avg. &\cellcolor{gray!10} - &\cellcolor{gray!10} - & \cellcolor{gray!10}\textbf{97.25} & \cellcolor{gray!10}\textbf{13.75} \\

\hline
\end{tabular}}

\end{table}

\begin{figure*}[t!]
\vspace{-10pt}
\centering    
\subfloat[Token] {
\includegraphics[width=0.3\textwidth]{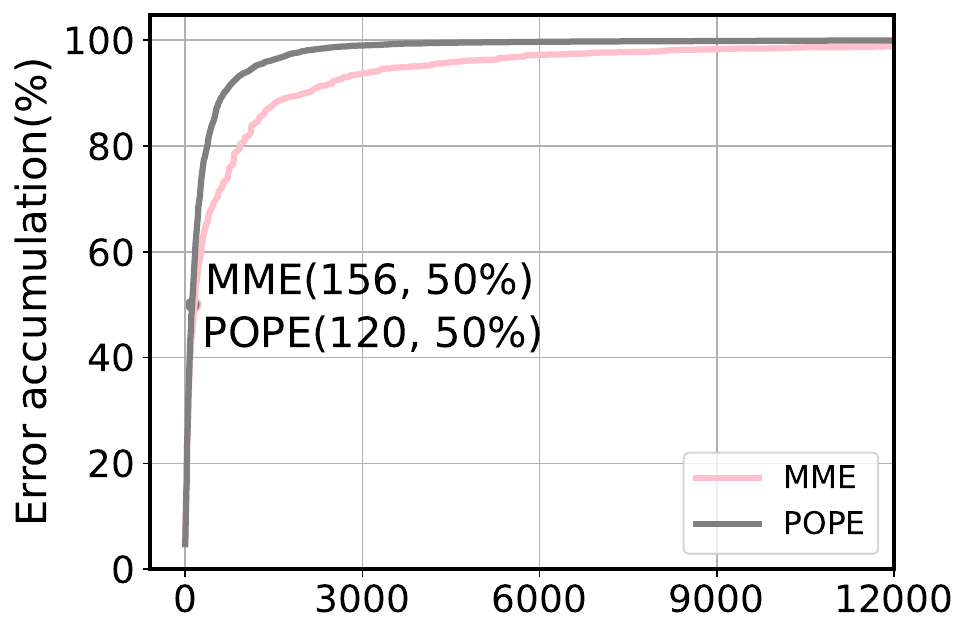}  
}     
\subfloat[Object] {  
\includegraphics[width=0.3\textwidth]{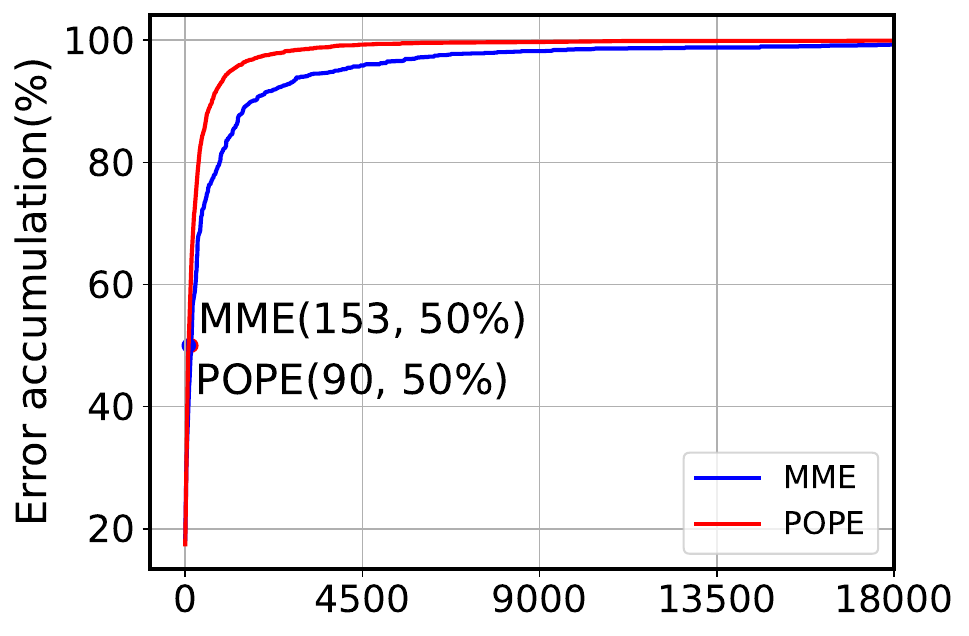} 
}
\subfloat[Co-occurrence] {  
\includegraphics[width=0.3\textwidth]{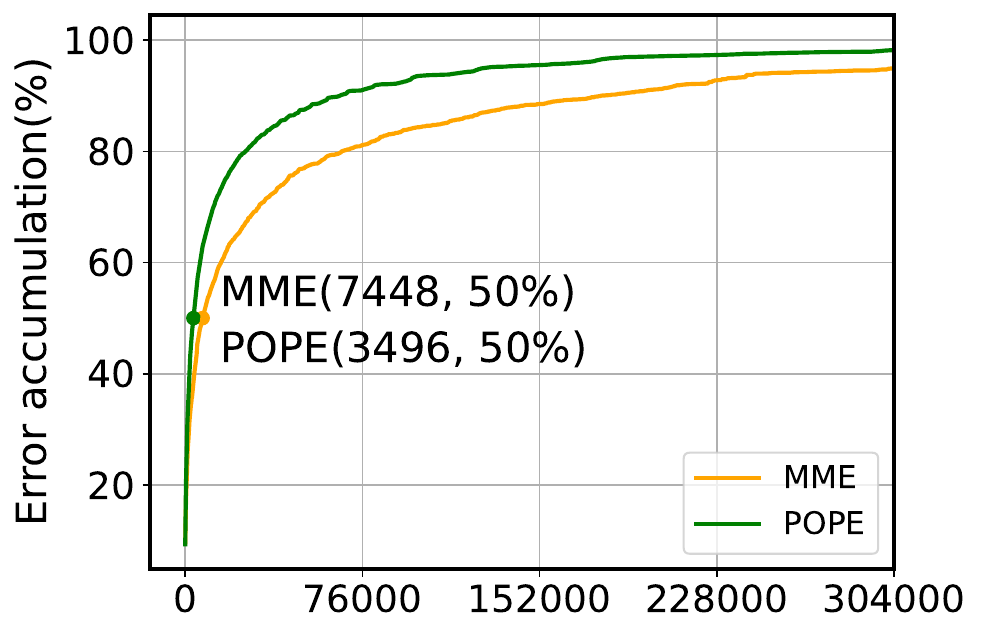}     
}
\vspace{-5pt}
\caption{
Error accumulation curve of POPE and MME based on the training data distribution. It reveals that tail entities contribute to the majority of failure cases.
(a) Token-level word distribution in MME \citep{fu2023MME} and POPE \citep{Li2023POPE}.
(b) Object-level word distribution in MME and POPE.
(c) Co-occurrence-level word distribution in MME and POPE.
}     
\label{fig:benchmark_wrong_distribution}     
\end{figure*}

\section{Analysis}
\subsection{Preliminary}
. 
In this paper, we focus primarily on the LT problem in the instruction-tuning process of LVLMs. 
In this paper, we primarily investigate the LT problem in the instruction-tuning process of LVLMs. During this process, the training data is typically structured as $D={(I, C)}$, where $I$ represents the image and $C$ denotes the corresponding conversation.

\begin{table}

\centering
\caption{Relative training data volume for the tail \textbf{50\%} of failed cases. ``Tok'', ``Obj'', and ``Co'' refer to Token, Object, and Co-occurrence, respectively. \textbf{\%E} denotes the percentage of tail entities, while \textbf{\%DI} represents the percentage of tail data instances.}
\label{tab:tail_more_error}
\resizebox{0.73\columnwidth}{!}{
\setlength{\extrarowheight}{-5pt}
\begin{tabular}{ccccc}
\toprule
\textbf{Data} & \textbf{Level} & \textbf{thres} & \textbf{\% E} & \textbf{\% DI}\\
\midrule
\multirow{3}{*}{\textbf{MME} \cite{fu2023MME}}         & Tok        & 156   & 99.98 & 75.23 \\
                                      & Obj       & 153   & 99.83 & 51.33 \\
                                      & Co     & 7448  & 99.02 & 69.62 \\
\midrule
\multirow{3}{*}{\textbf{POPE} \cite{Li2023POPE}}        & Tok        & 120   & 99.98 & 80.50 \\
                                      & Obj       & 90    & 99.71 & 59.76 \\
                                      & Co     & 3496  & 99.54 & 75.45 \\
\hline
\cellcolor{gray!10} Avg.& \cellcolor{gray!10} - & \cellcolor{gray!10} -  & \cellcolor{gray!10} \textbf{99.68} & \cellcolor{gray!10} \textbf{68.65} \\
\hline
\end{tabular}

}
\vspace{-10pt}
\end{table}

\subsection{Entity Distribution Construction}
\label{sec: entity distribution construction}
Specifically, we conduct the whole analysis procedure by constructing the frequency distribution of entities $Q_e$ from these four perspectives among the whole training set. The overall framework is as shown in Figure 2. Here we describe these four perspectives as below:

\paragraph{Token} entities are a set of meaningful nouns that are extracted from the text within data instances across the whole training set. $e_t = \{n| n \in \text{Noun} \land n \subseteq C \text{ for } (I,C)\text{ in } D \}$. Technically, we employ a Part of Speech (POS) parser to extract all nouns from each data instance within the training set, identifying them as token entities.

\vspace{-5pt}

\paragraph{Object} entities represent the objects that truly exist in the image within data instances. $e_o = \{o| o \in I \text{ for } (I,C)\text{ in } D \}$. We initially employ LLMs to extract all potential objects from the textual records of each data instance within the training set. The full prompts used to extract object information are detailed in the Appendix D.1. Subsequently, we input the image along with all token entities and LM-extracted objects into a visual grounding model, i.e., GroundingDINO \citep{liu2023groundingdino} to identify visual objects for each data instance, termed as object entities. Finally, we compute the frequency distribution of all object entities across the entire training set.

\vspace{-5pt}

\paragraph{Co-occurrence} entities refer to pairs of objects that appear together in the same image within a data instance. Formally, $e_c = \{(o_1,o_2) \| o_1 \in I \land o_2 \in I  \text{ for } (I,C)\text{ in } D \}$. Using the extracted object entities, we can construct a co-occurrence graph $G(V, E)$, where the vertex set $V$ consists of object entities and the edge set $E$ corresponds to the set of co-occurrence pairs $e_c$.

\vspace{-5pt}

\paragraph{Interrogation} entities are the questioning methods used in the text within data instances. $e_w = \{q| q  \in Q \land q \in C \text{ for } (I,C)\text{ in } D \}$. Where Q is the full question method set. We employ LLMs to extract all methods of posing questions from the data instances, defining them as interrogation entities. We extract all four kinds of entities from LLaVA \citep{liu2023improvedllava}'s instruction-tuning dataset.

\subsection{Reverse Indexing}
To study the severity of LT issues within the training data and build a connection between entities and data instances, we build a reverse indexing dictionary mapping from the entities in four different perspectives backward to the data instances. Subsequently, we use the number of data instances corresponding to each entity as frequency to build the \textbf{reversed distribution $Q_r$} of four perspectives, which can be formulated as $Q_r = \{ e_1: N_{e_1}, e_2: N_{e_2}, …, e_n: N_{e_n} \}$ where $e_i$ means entity item and $N_{e_i}$ means the number of corresponding data instances of $e_i$.

Taking LLaVA 1.5 \citep{liu2023improvedllava}'s instruction tuning data as an example, we count the number of data instance matches for each entity and build a reversed distribution based on the mapping data. The thresholds of the tail data and relative data volume are shown in Table \ref{tab:reverse-indexing}. 
Surprisingly, among four perspectives, an average of \textbf{97.25\%} entries account for only \textbf{13.75\%} data instances on average, which can partially illustrate the scarcity of tail data and severity of the long-tail problem existing in training data of LVLMs.

\subsection{Significance of Long-Tail Problem Mitigation}

\textbf{Tail data accounts for more failed cases}. 
We first evaluate LLaVA 1.5 on two popular benchmarks, POPE and MME, and analyze the failed cases, as incorrect responses generated by LVLMs often reveal the model’s weaknesses. 
Subsequently, we rank the failed cases according to the entity distribution of LLaVA’s instruction-tuning data and extract the bottom (tail) \textbf{50\%} of these cases. 

Next, we measure the number of entities and data instances corresponding to these errors. The results, presented in Table \ref{tab:tail_more_error}, reveal that the tail 50\% of failed cases cover over \textbf{99\%} of entities and account for an average of \textbf{68.65\%} of training instances. 
Additionally, we present a cumulative error curve with entity distribution on the horizontal axis, ordered from most to least frequent. 
{As shown in Figure \ref{fig:benchmark_wrong_distribution}, it can be observed that tail entities account for the majority of failure cases. Moreover, the distribution location of failed cases is positioned further towards the tail of the distribution compared to correct answers. The detailed results of location analysis are shown in Appendix B.2.}

\noindent\textbf{Distribution varies between train and test data.} Besides, the distribution of train and test data is also different. In statistics-based deep learning, it is assumed that the training data maintains the same distribution as the evaluation data. Based on the intuition that a larger bias between distributions can result in performance loss, we examine the differences between the entity distributions of the training and evaluation data. We select the training data of LLaVA 1.5 \citep{liu2023improvedllava}, as well as the evaluation data from POPE \citep{Li2023POPE} and MME \citep{fu2023MME}. 
The resulting co-distribution is presented in Figure \ref{fig:long_tail_distribution_analysis}. Notably, a clear difference between the distributions of the evaluation and training data can be observed.

Therefore, addressing LT issues in the training data is crucial for LVLMs to enhance their understanding of underrepresented concepts and improve overall performance.

%% file: Styles/content/3.approach.tex
\section{Approach}

\subsection{Data Rebalancing Stage}

 To mitigate the long-tail problem in the LVLM, our adaptive data refinement framework starts by alleviating the redundancy problem existing in training data. Concretely, we achieve this by flattening the exponential distribution and decimating the duplicated entities.

\begin{table*}[t!]
\caption{\textbf{Comparison with models trained with different methods on different benchmarks.} IT represents the number of training instances used during instruction tuning. \textcolor{blue}{+DR} denotes the results after the data rebalancing stage, while \textcolor{red}{+DS} represents the results following the data synthesis stage. Benchmark names are abbreviated due to space limits. *: ShareGPT4V's instruction tuning stage refers to the 2nd stage (3 in total). The best results are indicated in \textbf{bold}.}

\label{tab:main_results}
\centering
\resizebox{\textwidth}{!}{
\begin{tabular}{lc|ccccccccccc }
\toprule
Method & IT* & VQA$^\text{OK}$ & SEED$^{\text{2}}$ & QB$^\text{2}$ & MMS & MME$^\text{P}$ & SQA$^\text{I}$ & MMMU & VQA$^\text{T}$ & GQA & MMB & VQA$^\text{v2}$ \\
\midrule
LLaVA 1.5 & 665.0K & 53.2 & 48.7 & 47.3 & 33.5 & 1510.7 & 69.3 & 35.3 & 46.0 & 61.9 & 64.3 & 76.6\\ 
\hspace{0.3cm}\textcolor{blue}{+DR} & 581.0K & 55.3 & 57.2 & 46.8 & 33.8 & 1470.6 & 69.5 & 34.8 & 46.0 & 62.8 & \textbf{65.5} & 76.9 \\ 
\hspace{0.3cm}\textcolor{blue}{+DR} \textcolor{red}{+DS} & 665.0K & \textbf{57.4} & \textbf{57.4} & \textbf{49.6} & \textbf{35.5} & \textbf{1512.8} & \textbf{70.4} & \textbf{36.7} & \textbf{47.2} & \textbf{62.9} & 65.0 & \textbf{76.9} \\ 
\hline
ShareGPT4V & 1246.0K & 54.0 & 59.6 & 44.2 & 34.7 & 1560.4 & 68.9 & 35.1 & 50.2 & 63.3 & 68.0 & 78.6 \\ 
\hspace{0.3cm}\textcolor{blue}{+DR} & 1168.0K & 56.7 & 59.6 & 44.9 & 35.0 & 1542.3 & 68.6 & 35.7 & \textbf{50.9} & \textbf{63.9} & 67.9 & 78.7 \\ 
\hspace{0.3cm}\textcolor{blue}{+DR} \textcolor{red}{+DS} & 1246.0K & \textbf{57.9} & \textbf{59.9} & \textbf{45.7} & \textbf{35.5} & \textbf{1564.9} & \textbf{69.4} & \textbf{36.1} & 50.9 & 63.7 & \textbf{68.8} & \textbf{78.7} \\ 

\bottomrule
\end{tabular}
}
\end{table*}

\subsubsection{Probability Dictionary Construction}
DR stage starts with settling down the resampling ratios of redundant data instances.
First, we use the entity distribution construction method mentioned in Section \ref{sec: entity distribution construction} to construct the distribution dictionary $Q_e$ for each entity within all selected perspectives $C$. Subsequently, we construct the reverse indexed dictionary and use the number of data instances mapped with entities $N_e$ as frequency to build a reversed distribution $Q_r$, which is used for calculating the sampling ratio $p_s$. A threshold $\tau$ is used to distinguish the head and tail data. $\tau$ is an entity's position in $Q_r$ while entities before $\tau$ count for a small ratio of all entities but are mapped to massive data instances. We set $\tau$ as indicated in Table \ref{tab:reverse-indexing}, consistent with the values used in the analysis stage. For an entity $e$ of perspective $x$, we set the probability of sampling by $P_{e_x} = \tau_x/N_{e_x}$.

After constructing the probability dictionary of each entity $e$, we start from $Q_e$ to sample the selected data. Since each data instance contains several entities in $Q_e$, we sample every entity among one instance. So we introduce a new hyperparameter $n_p$, which means the data instance with the total number of sampled entities over $n_p$ is selected. We conduct this procedure over the full dataset, and the final selected core set is denoted as $D^*$. 
The detailed method is demonstrated in Appendix C.

\subsection{Data Synthesis Stage}   
Despite the presence of redundancy in the head entities, the issue of scarcity in the tail entities still persists. To alleviate the issue of scarcity, we design the data synthesis methods from the perspective of vision and language. Figure \ref{fig:main_figure} displays the full data adjusting framework.

\subsubsection{Language Data Synthesis}
\label{sec:language_data_augment}
The core idea of our DS stage is to replace head concepts with tail ones. First, we use WordNet \citep{fellbaum1998wordnet} to extract all synonyms of token entities and construct a mapping system. For each head instance, we extract the linguistic entities, search for their synonyms in the mapper, and filter out the head ones. Next, we feed the original head conversation into a language model (LM) and prompt the LLMs to rewrite the conversation using the selected tail synonyms. Full prompts to instruct LLMs can be found in Appendix D.2. It is important to note that certain stop words would not be replaced.

\begin{table}[t!]
\caption{Performance comparison across existing data balancing methods. The best results are indicated in \textbf{bold}, and the second-best results are \underline{underlined}. IT represents the number of training instances used during instruction tuning.}
\label{tab:compare_balance}
\centering
\resizebox{\columnwidth}{!}{

\begin{tabular}{lc|ccccc }
\toprule
Method & IT & GQA  & SEED & SEED$^\text{v2}$ & POPE & MMB  \\
\midrule
Baseline & 665K & 62.0 & 61.0 & 57.2 & 87.2 & 65.5 \\
EL2N & 581K & 62.5 & 53.6 & 47.4 & 87.2 & 65.2\\
Perplexity & 581K & 62.3 & 53.4 & 47.4 & 86.8 & 63.7\\
CLIP Score & 581K & 62.5 & 53.0 & 47.0 & 87.0 & 64.5\\
COINCIDE & 133K & 59.8 & - & - & 86.1 & 63.1\\
\hline
\cellcolor{gray!10}Ours-DR & \cellcolor{gray!10} 581K &  \cellcolor{gray!10}\underline{62.8} &\cellcolor{gray!10} \underline{61.0} & \cellcolor{gray!10}\underline{57.2} & \cellcolor{gray!10}\underline{87.2} & \cellcolor{gray!10}\textbf{65.5}\\
\cellcolor{gray!10}Ours & \cellcolor{gray!10}665K & \cellcolor{gray!10}\textbf{62.9} &\cellcolor{gray!10} \textbf{61.3} & \cellcolor{gray!10}\textbf{57.4} & \cellcolor{gray!10}\textbf{87.4} & \cellcolor{gray!10}\underline{65.0} \\
\hline

\end{tabular}
}
\vspace{-10pt}

\end{table}

\subsubsection{Diffusion-Based Visual Data Synthesis}
\label{sec:diff_data_augment}
In addition to the language synthesis method, we propose a more comprehensive multiple-perspective data synthesis approach. Editing tail images into different styles without altering the key entities can address the scarcity of objects and co-occurring objects. Meanwhile, a rewriting process can resolve the scarcity of tokens and interrogation methods. In our analysis of the LT problem, we determine whether a data instance is selected using $P_e$ and $n_p$. 

However, in certain cases, the probability $P_e$ may exceed 1. The absolute value of $P_e$ also provides some insight into the scarcity of a data instance. This value can be used to decide how many instances to synthesize. Notably, \textbf{76\%}  of the synthetic data has a $P_e$ value of \textbf{less than 5}, so we utilize this method to determine the synthetic quantity for each data instance. The synthetic quantity for each data instance $d=(I,C)$ can be calculated as:
\vspace{-5pt}
\begin{equation}
P_d^* = \max_{e,x}{P_{e_x}} ; N_{d,aug} = 
\begin{cases} 
0  & \text{if }  P_d^* < 1, \\
\left\lfloor \sqrt{P_d^*} \right\rfloor & \text{if } 1 \leq P_d^* < 5, \\
2  & \text{if } 5 \leq P_d^*, 
\end{cases}
\label{eq:aug_num}
\end{equation}
\vspace{-5pt}

Given a tail instance $d_t=(I_t, C_t)$, our objective is to generate an image similar to $I$ and produce corresponding instruction data, specifically in the form of conversations. To achieve the goal of subtly altering an image’s style while retaining its key information (i.e. objects or their co-occurrence), we propose two methods, both built upon a diffusion-based model $G$ \citep{croitoru2023diffusion_servey_1,ho2020ddpm_paper20,ddpm4image_syn}.

We leverage ControlNet \cite{zhang2023controlnet} as our DDPM model, which can generate a new image based on an existing image $I$ and a natural language prompt $t$. The generating procedure of ControlNet is represented as $I_{t,\text{syn}} = G(I_t, p_{\text{def}})$ where $p_{\text{def}}$ denotes the default prompt used to preserve the primary information of the input image. Through a detailed comparison of generation quality, we found that compared to traditional DDPM, using ControlNet is more effective in preserving image details and key objects.

With the generated image $I_{t,\text{syn}}$, we can obtain a descriptive conversation paired with the image to serve as the instruction tuning data instance. We utilize an off-the-shelf vision captioner to generate captions $Cap_{t,\text{syn}}$ for \(I_{t,\text{syn}}\). Subsequently, we extend the captions into conversations, as required by visual instruction tuning. During this stage, we use LLMs to expand the captions into full conversations, with the prompts for expansion provided in Appendix D.2. This process enables us to effectively synthesize scarce data instances, helping to alleviate the LT problem from all four perspectives. Finally, we augment the rebalanced dataset with synthetic data to restore its scale to match that of the LLaVA base model.

Given the generated image $I_{t,\text{syn}}$, we derive a descriptive conversation paired with the image, forming an instruction-tuning data instance. Specifically, we employ an off-the-shelf vision captioner to produce the relative captions $Cap_{t,\text{syn}}$ for $I_{t,\text{syn}}$. These captions are then expanded into conversations, as visual instruction tuning requires. For this step, we utilize LLMs to transform the captions into comprehensive conversations, following the prompts detailed in Appendix D.2. This methodology allows us to effectively synthesize scarce data instances, mitigating the LT problem from all four perspectives. Finally, we augment the rebalanced dataset with synthetic data to match the original data scale of the LLaVA base model.

%% file: Styles/content/4.experiment.tex
\section{Experiments}
\subsection{Baseline Models}
In this paper, we use LLaVA 1.5 and ShareGPT4V as baselines. We apply ADR to their instruction tuning data and train our model on the adjusted data to verify its effectiveness. Below is an overview of the two models:
\begin{itemize}[leftmargin=*]
\setlength{\itemsep}{0pt}

\item \textbf{LLaVA 1.5} \citep{liu2024visual,liu2023improvedllava} represents a novel end-to-end trained large multimodal model that combines a vision encoder and Vicuna for general-purpose visual and language understanding, achieving impressive chat capabilities.

\item \textbf{ShareGPT4V} \citep{chen2023sharegpt4v} uses the adjusted training data obtained by GPT-4 and post-trained ShareCapioner and improves the performance of existing VLMs. Though they focus on data adjustment either, the long-tail problem remains, that is, our method is orthogonal to ShareGPT4V.
\end{itemize}

\subsection{Benchmarks}
We utilize a comprehensive set of widely recognized benchmarks, spanning a broad range of academic VQA tasks and recent benchmarks designed to test the extensive abilities of LVLMs. The VQA series benchmarks \citep{goyal2017vqav2,marino2019okvqa,singh2019textvqa} represent traditional, comprehensive VQA tasks. GQA \citep{hudson2019gqa} evaluates multiple reasoning skills and spatial understanding, which presents a greater challenge. The MME Benchmark \citep{fu2023MME} assesses the comprehensive capabilities of LVLMs through a series of carefully crafted questions spanning 14 distinct sub-tasks. MMBench and MMBench-CN \citep{MMBench} are designed to assess the model’s vision-based reasoning and perceptual abilities in both English and Chinese. MMSTAR \citep{chen2024mmstar} and SEED \citep{li2023seed} evaluate the model's comprehensive ability from different aspects. ScienceQA \citep{lu2022learn} evaluates LVLMs on multimodal, multiple-choice science questions, while MMMU \citep{yue2023mmmu} tests LVLMs across multiple disciplines, requiring college-level subject knowledge and sophisticated reasoning. Finally, Q-Bench \citep{wu2024qbench} focuses on assessing low-level perception. All benchmarks we used and their abbreviations can be found in Appendix A.1.

\begin{figure}[!t]
    \centering
    \includegraphics[width=0.9\linewidth]{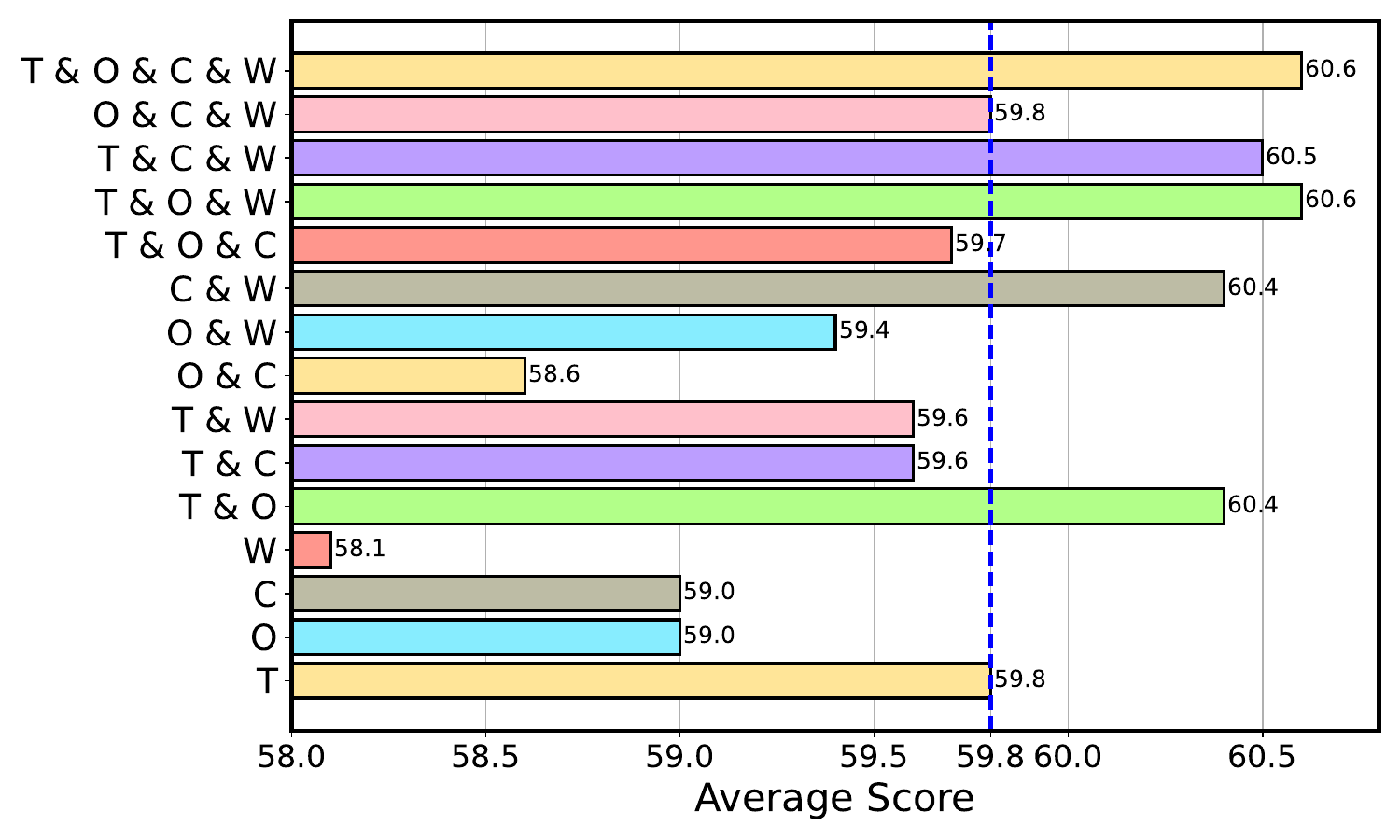}
    \caption{
            Ablation study on data rebalancing combinations. T, O, C, and W refer to Token, Object, Co-occurrence, and Interrogation respectively. The values displayed in the graph represent average scores across a variety of comprehensive benchmarks. The blue dashed line indicates the baseline performance of LLaVA 1.5.
    }
    \label{fig:ablations_rebalance}
\end{figure}

\subsection{Results For Comprehensive Evaluation}

\begin{table}
\centering
\caption{Tail concept prediction accuracy (\%) on
 ScienceQA-IMG \citep{lu2022learn} dataset. Tail@$k\%$ (simplified as @$k$), head@$k\%$ (simplified as H@$k$), and overall accuracy are reported. \textcolor{blue}{+DR} denotes the results after data rebalancing, while \textcolor{red}{+DS} represents the results following the data synthesis stage. \textbf{Bold} numbers represent the best results across all methods.  IT represents the number of training instances used during instruction tuning.}
 \label{tab:tail_concept_acc}
\resizebox{0.95\columnwidth}{!}{
\begin{tabular}{lc cccccccc }
\toprule
\multirow{2}{*}[-0.66ex]{Methods} & \multirow{2}{*}[-0.66ex]{IT}& \multicolumn{6}{c}{ScienceQA} \\
\cmidrule(lr){3-8}
 && @5   &@10  & @15  &  @20  & H@80   &Overall \\ 
\midrule
LLaVA 1.5 & 665.0K & 67.9 & 70.0 & 67.9 & 68.5 &74.6 &69.3  \\
\hspace{0.3cm}\textcolor{blue}{+DR} & 581.0K & 69.2& 69.7 & 67.8 &68.5 &76.2 &69.5  \\ 
\hspace{0.3cm}\textcolor{blue}{+DR} \textcolor{red}{+DS} & 665.0K & \textbf{70.1} &\textbf{70.5} & \textbf{68.3} &\textbf{69.0} & \textbf{78.6} &\textbf{70.2} \\ 

\bottomrule
\end{tabular}
}

\end{table}

The results on the eleven selected benchmarks are shown in Table \ref{tab:main_results}. By balancing the training data while retaining 87\% of its original scale, our method outperforms the baseline on most benchmarks. After the DS stage, we restore the same data scale as LLaVA 1.5. Without incorporating additional data, our method achieves an average absolute improvement of 2.28 points and a relative improvement of \textbf{4.36\%}. Notably, on challenging benchmarks such as MMStar, SEED Bench2, OK-VQA, and Qbench-2, it surpasses the baseline with an average absolute improvement of 4.3 points and a relative improvement of \textbf{9.15\%}.

\renewcommand{\thefootnote}{\arabic{footnote}}

Moreover, we display the comparison between our method and popular data-balancing methods such as EL2N \cite{paul2021diet_el2n}, Perplexity \citep{marion2023whenlessismore_perplexity}, CLIP Score, and COINCIDE \citep{lee2024concept}\footnote[1]{COINCIDE did not release code, we compare results from their paper.} in Table \ref{tab:compare_balance}. Our method consistently outperforms the baselines across the majority of benchmarks, which cover a wide range of comprehensive tasks. Additionally, our approach focuses on mitigating LT issues and is orthogonal to most existing data balancing and augmentation methods. This means it can be applied alongside those techniques to achieve even better performance across various benchmarks. The detailed experimental results are provided in Appendix A.

\begin{figure}[!t]
    \centering
    \includegraphics[width=0.9\linewidth]{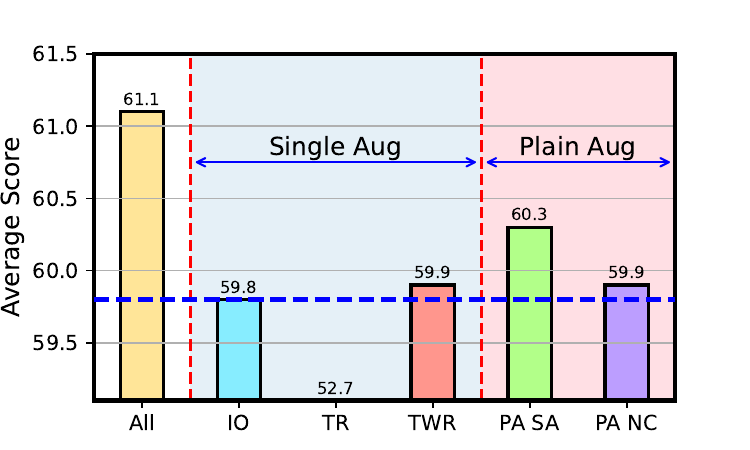}

    \caption{
        Ablation study on data synthesis methods. The meaning of abbrs occurred here is explained in Section \ref{sec:ablationofaug}. The values displayed in the graph represent average scores across a variety of comprehensive benchmarks. The blue dashed line indicates the baseline performance of LLaVA 1.5.
    }
    \label{fig:ablations_aug}
\end{figure}

\subsection{Performance on Tail Instances}

In addition to the main results on comprehensive benchmarks, we assess the model’s performance on tail concepts to validate the effectiveness of improving tail concept performance. 
We selected ScienceQA \citep{lu2022learn} and VQAV2 \citep{goyal2017vqav2} to evaluate performance on tail data. First, we applied the same method described in Section \ref{sec: entity distribution construction} to extract entities from the selected benchmark data and build their reverse indexed distribution. Subsequently, for each data instance, we calculated the average distribution position across each perspective and determined whether the data instance falls into the tail category by $ \mathbbm{1}(average(L_i)>\tau_R) $. Here, \(\mathbbm{1}\) represents an indicator function. We set different thresholds to make sure to get different ratios of tail data. We then extract the tail data instances of different rations and evaluate the performance accordingly. The complete results for tail performance are presented in Table \ref{tab:tail_concept_acc} and Figure \ref{fig:head_picture}(b). As shown in the table, our method effectively improves tail performance \textbf{without compromising the head or overall performance}, demonstrating the efficacy of our approach.

%% file: Styles/content/5.ablation.tex
\section{Ablation Study}

\subsection{Different Combinations of Perspectives}
To determine the most effective rebalancing and synthesis method, we train the model using data processed with different combinations of perspectives and subsequently evaluate the target model. The results are presented in Figure \ref{fig:ablations_rebalance}. We assess these combinations based on average performance across different benchmarks, the number of top-ranked results, and performance stability. While several combinations achieved the highest average performance, the full combination of all perspectives proved to be the most stable, as it ranked first in both the number of top performances and stability. Detailed results can be found in the Appendix A.2.

\subsection{Synthesis Methods}
\label{sec:ablationofaug}
In addition to the perspective combination selection during the Data Rebalancing stage, we present our ablation study on different synthesis methods. Synthesis is performed on the same rebalanced data checkpoint across all perspectives to determine which method is the most effective.

We select six synthesis methods to compare, divided into language synthesis methods (Sec.\ref{sec:language_data_augment}) and vision synthesis methods (Sec.\ref{sec:diff_data_augment}). The following methods are tested:
\begin{itemize}[leftmargin=*]
\setlength{\itemsep}{0pt}

\item \textbf{All}: Tail instances are selected from all four perspectives, with full visual data synthesis (Sec.\ref{sec:diff_data_augment}) applied.

\item \textbf{Image Only (IO)}: Tail instances are selected from all four perspectives, applying visual data synthesis (Sec.\ref{sec:diff_data_augment}), but the conversation text remains unchanged.

\item \textbf{Token Rewrite (TR)}: Full language data synthesis (Sec.\ref{sec:language_data_augment}) methods are applied.

\item \textbf{TW Rewrite (TWR)}: Tail instances are selected based on Token and Interrogation perspectives, and the conversations are rewritten using a language model (LM).

\item \textbf{PlainAug SimpAdd (PA SA)}: Tail data are selected from all four perspectives, and simple resampling is applied.

\item \textbf{PainAug NewCap (PA NC)}: Tail data are selected from all four perspectives, followed by re-captioning, with the new captions incorporated into conversations using the same method with Sec.\ref{sec:diff_data_augment}.

\end{itemize}

We use these synthesis methods to restore the data to 665K and test which checkpoint yields the best performance using the same volume of training data. We assess these methods based on average performance across different comprehensive benchmarks. The results are displayed in Figure \ref{fig:ablations_aug}, while detailed results can be found in the Appendix A.2.

%% file: Styles/content/6.conclusion.tex
\begin{figure}[!t]
    \centering
    \includegraphics[width=\linewidth]{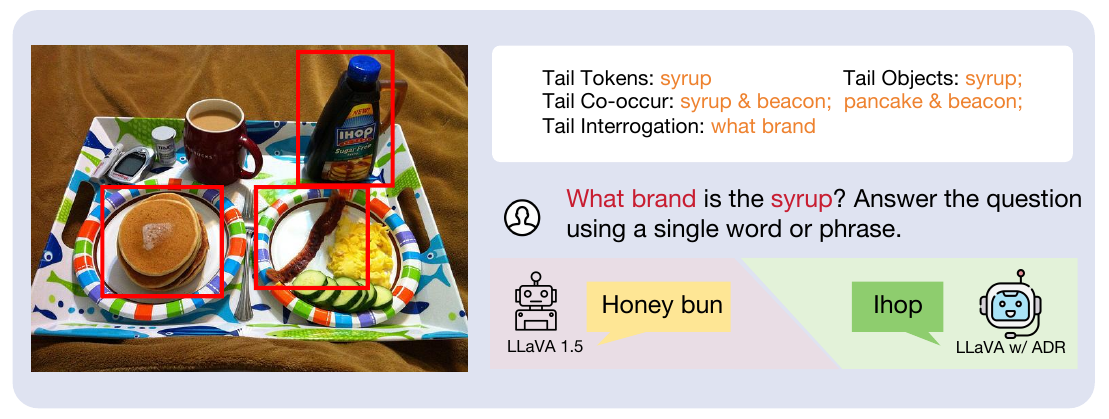}
    \caption{
   Qualitative comparison between the baseline model (LLaVA 1.5) and our proposed method (LLaVA w/ ADR) on a tail example. LLaVA w/ ADR can handle tail questions while LLaVA 1.5 fails to answer. While LLaVA 1.5 fails to answer tail questions, LLaVA w/ ADR successfully addresses them.
    }
    \label{fig:quality_1}
\end{figure}

\begin{figure}[!t]
    \centering
    \includegraphics[width=\linewidth]{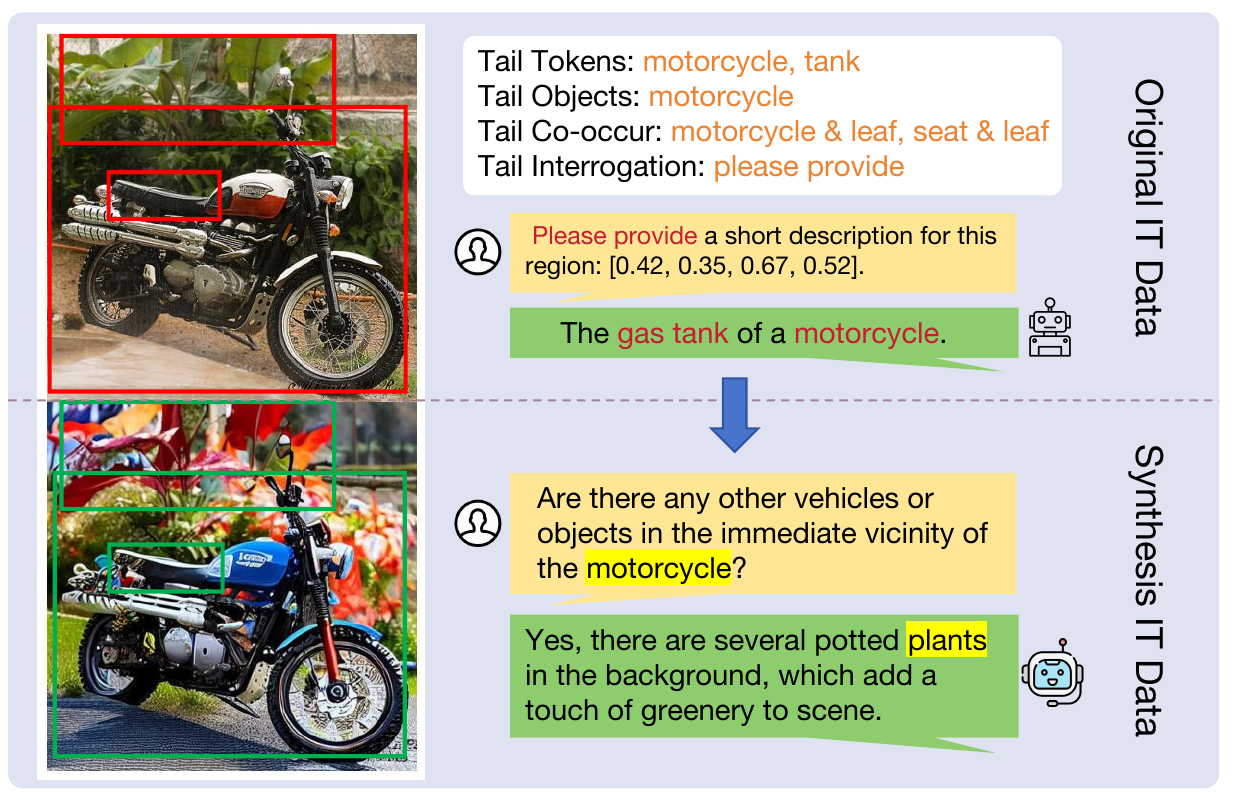}
    \caption{
Comparison between the original instruction-tuning (IT) data and our synthesized IT data. Tail concepts in the original data are highlighted using \red{red} boxes and fonts, whereas synthesized tail concepts are marked with \textcolor{green}{green} boxes and \colorbox{yellow}{yellow} fonts.
    }
    \label{fig:synthesis_1}
\end{figure}

\section{Qualitive Results}
\subsection{Comparison on Tail Examples}

We compare our method (ADR) and baseline on several tail cases, the results are demonstrated in Figure \ref{fig:quality_1}, Where the baseline model fails to provide the correct answer, while our method (ADR) successfully handles the case. This highlights the ability of ADR to generalize better on challenging, less frequent instances, showcasing its robustness in scenarios where baseline models often struggle.

\subsection{Data Synthesis Results}

Figure \ref{fig:synthesis_1} provides a direct example of our synthesized data. The synthesis process starts by generating an image that closely mirrors those containing tail concepts, which are typically underrepresented in conventional datasets. This step ensures that the visual features of these infrequent concepts are well captured. Subsequently, we generate corresponding textual data, such as descriptive captions or questions, carefully designed to complement the visual content and enhance the contextual understanding of the tail concepts.

As illustrated in Figure \ref{fig:synthesis_1}, our synthesis approach not only produces high-quality visual content but also enriches the associated textual representation. This dual enhancement across both visual and textual modalities effectively addresses the data imbalance, significantly improving the representation of tail concepts. Consequently, our method effectively enhances model generalization and performance in underrepresented scenarios.

\section{Conclusion}
In this paper, we study the long-tail problem existing in the instruction tuning data of LVLM from four perspectives and make the first attempt to mitigate it. 
Our analysis reveals that unbalanced data can result in a performance gap between the head data and the tail data, therefore harming the model's performance. Based on this insight, we develop an Adaptive Data Refinement Framework (ADR), which first balances the redundant head data and then adjusts the unbalanced tail distribution. Experimental results demonstrate that our method improves the tail performance and overall performance without harming the head part performance. 

In the future, we plan to extend our method to the pre-training (PT) stage of LVLMs.

\section{Acknowledgement}

This work was supported by the Shanghai Artificial Intelligence Laboratory. We sincerely appreciate their support and resources, which contributed to the successful completion of this research.

%% file: Styles/content/8.5.new_appendix.tex
\appendix
\section{Details of Experiments}
\label{appendix:experiment}

\subsection{Benchmarks}
\label{appendix:benchmarks}

All benchmarks we used and their abbreviations are introduced as follows. 

\setlength{\itemsep}{0pt}
\begin{itemize}[leftmargin=*]
\setlength{\itemsep}{0pt}

 \item \textbf{VQA$^\text{v2}$}: The Visual Question Answering v2 dataset~\citep{goyal2017vqav2} consists of 265,016 images, each with 5.4 questions on average, requiring vision, language, and commonsense understanding, with 10 ground-truth and 3 plausible but incorrect answers for evaluation.

 \item \textbf{VQA$^\text{T}$}: The TextVQA dataset~\citep{singh2019textvqa} includes 45,336 questions over 28,408 OpenImages images, requiring models to read and reason about text within images for answers.

 \item \textbf{VQA$^\text{OK}$}: The Open-Ended Knowledge Visual Question Answering dataset~\citep{marino2019okvqa} includes over 14,000 questions that require integrating visual content with external knowledge, such as Wikipedia, for final accurate answers.

 \item \textbf{GQA}: GQA~\citep{hudson2019gqa} is a large-scale dataset comprising over 22 million questions generated from scene graphs of 113,000 images. It is specifically designed to assess models on visual reasoning and compositional question answering, with a focus on reducing language biases.
 
 \item \textbf{SQA$^\text{I}$}: ScienceQA-IMG~\citep{lu2022learn} is a multimodal dataset comprising 21,208 science questions, each accompanied by corresponding images and explanations. It is designed to evaluate models’ capabilities in answering science-related questions through multimodal reasoning.

 \item \textbf{POPE}: The Polling-based Object Probing benchmark~\citep{Li2023POPE} evaluates vision-language models’ ability to detect hallucination by prompting them with classification questions regarding the presence of specific objects in an image.

 \item \textbf{SEED}: The SEED Bench~\citep{li2023seed} is a large-scale benchmark with 19,000 multiple-choice questions across 12 dimensions, designed for efficient evaluation of LVLMs without human intervention.

 \item \textbf{SEED$^\text{2}$}: SEED Bench v2~\citep{li2023seed2} is a comprehensive benchmark with 24,000 multiple-choice questions across 27 dimensions, comprehensively evaluating text and image generation capabilities of LVLMs.

 \item \textbf{MMMU}: The Massive Multi-discipline Multimodal Understanding benchmark \citep{yue2023mmmu} is designed to evaluate multimodal models on complex, college-level tasks that require subject-specific knowledge and advanced reasoning. 
 
 \item \textbf{MME$^\text{P}$}: The Multimodal Evaluation Benchmark~\citep{fu2023MME} assesses LVLMs’ perception and cognition through 14 subtasks, including object recognition and reasoning. This paper focuses on its perception subset.
 
 \item \textbf{MMB$^\text{CN}$}: MMBench~\citep{MMBench} is a benchmark with 3,000 multiple-choice questions across 20 dimensions, assessing vision-language models’ perceptual and cognitive abilities. CN denotes its Chinese validation set.
 
 \item \textbf{MMB}: The English validation subset of MMBench~\citep{MMBench}; 
 
 \item \textbf{MMS}: MMStar~\citep{chen2024mmstar} is a benchmark with 1,500 samples, assessing six core capabilities across 18 axes to evaluate LVLMs’ visual comprehension in complex scenarios.
 
 \item \textbf{QB$^\text{2}$}: Q-Bench 2~\citep{wu2024qbench} is a benchmark for evaluating multi-modal models on low-level vision tasks, focusing on visual perception, description, and quality assessment with datasets like LLVisionQA and LLDescribe.

\end{itemize}

\subsection{Detailed Results of Ablation Study}
\label{appendix:supply_ablation}
We conducted an ablation study on different balancing combinations and synthesis methods. In the ablation study of different rebalancing combinations, we conduct the DR stage using different combinations of four perspectives, i.e., one or more from (Token, Object, Co-occurrence, and Interrogation) to validate the effectiveness of different perspectives. The detailed results of the balancing ablation experiment are presented in Table \ref{tab:ablation_different_combination}.  Although some checkpoints achieved similar average results, we found that combining all perspectives yields the best performance in terms of both the number of top results and performance stability. 

Additionally, we conducted an ablation study on different synthesis methods. The results of the augmentation and synthesis experiments are presented in Table \ref{tab:ablation_different_augment}. Obviously, synthesizing from \textbf{ALL} perspectives (as outlined in Section 4.2.2) yields the best performance.

\subsection{Detailed Results of Main Experiment}
Beyond the main experiments, we conduct pure data augmentation on the original instruction-tuning dataset of LLaVA 1.5, focusing solely on the DS stage applied to the original training data. The resulting augmented data is used to instruction-tune LLaVA, which is then evaluated on various benchmarks. As shown in Table \ref{tab:detailed_main_results}, our ADR framework consistently surpasses most pure augmentation checkpoints on the majority of benchmarks, with a few exceptions, such as MMMU, MMB, and VQA$^\text{v2}$.

\subsection{Qualitive Results}
\begin{figure}[t]
\centering    

\begin{subfigure}[b]{\columnwidth}
    \centering
    \includegraphics[width=\columnwidth]{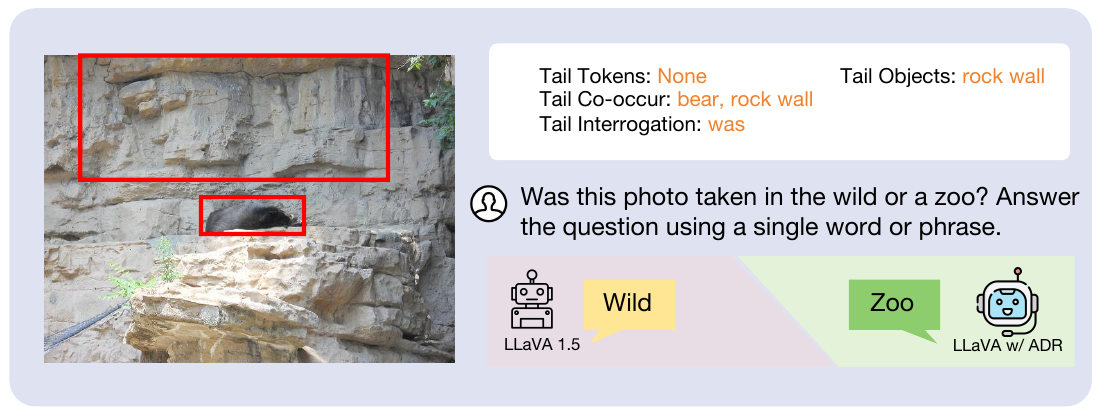}
    \caption{A bear resting peacefully beside a rock wall.}
\end{subfigure}

\begin{subfigure}[b]{\columnwidth}
    \centering
    \includegraphics[width=\columnwidth]{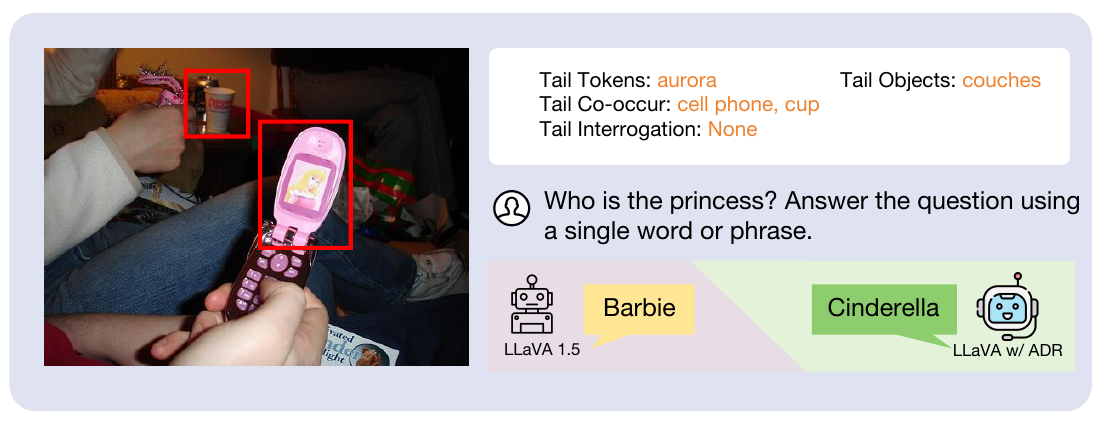}
    \caption{A cell phone displaying a cartoon princess on its screen.}
\end{subfigure}

\begin{subfigure}[b]{\columnwidth}
    \centering
    \includegraphics[width=\columnwidth]{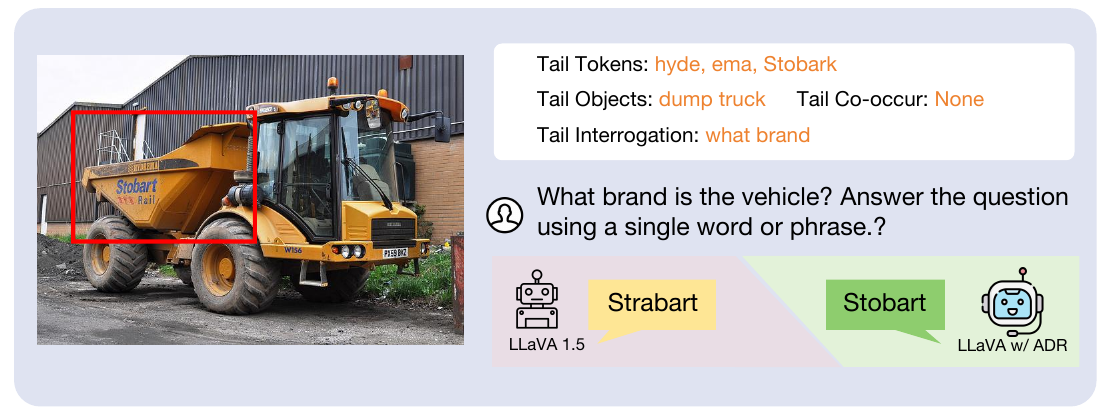}
    \caption{A dump truck.}
\end{subfigure}

\caption{
Qualitative comparison between the baseline model (LLaVA 1.5) and our proposed method (LLaVA w/ ADR) on a few tail examples. While LLaVA 1.5 fails to answer tail questions, LLaVA w/ ADR successfully addresses them.
}
\label{fig:appendix_qualitiveres} 

\end{figure}

\begin{figure}[t]
\centering    

\begin{subfigure}[b]{\columnwidth}
    \centering
    \includegraphics[width=\columnwidth]{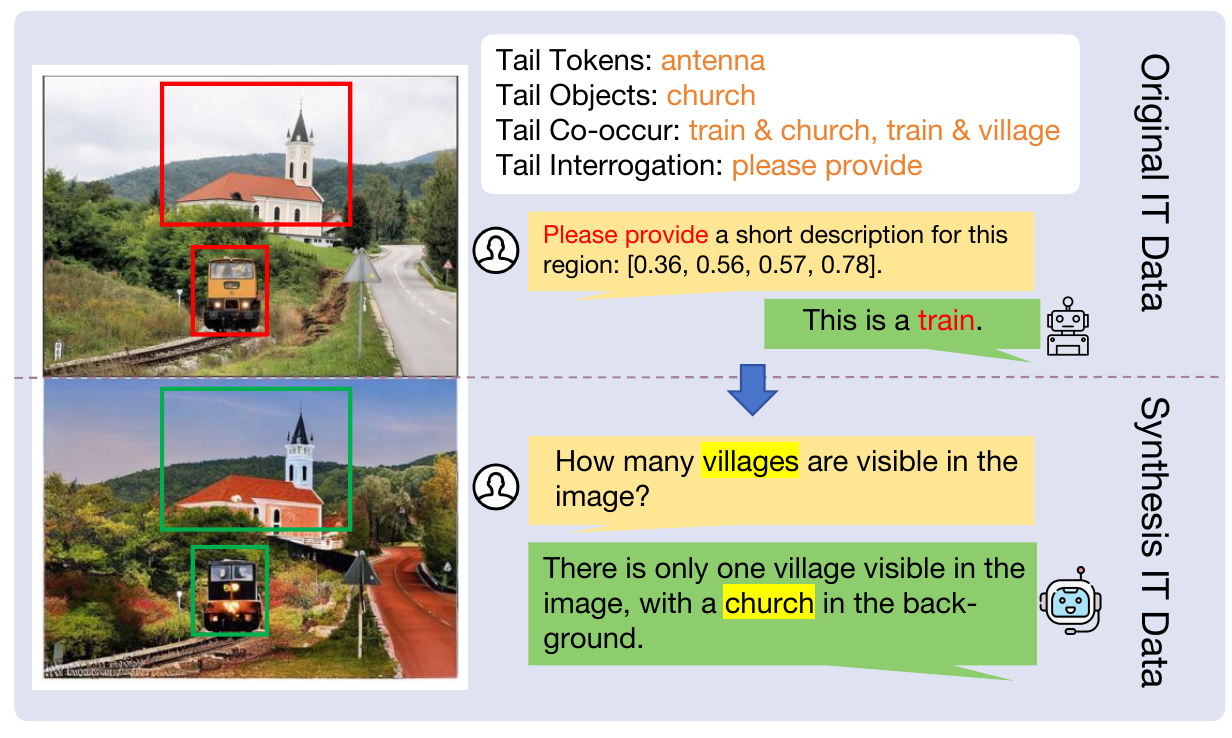}
    \caption{A train traveling along a railway near a church.}
\end{subfigure}

\begin{subfigure}[b]{\columnwidth}
    \centering
    \includegraphics[width=\columnwidth]{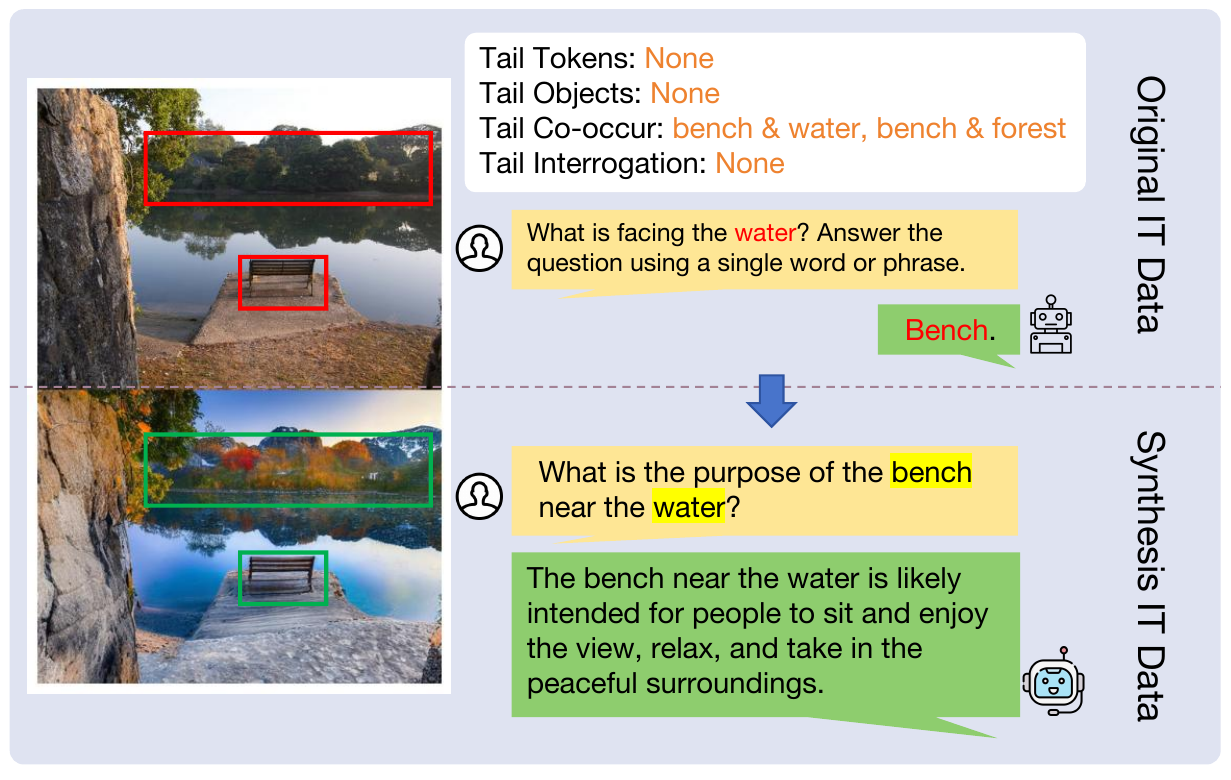}
    \caption{A bench by the lake, with a forest on the opposite shore.}
\end{subfigure}

\begin{subfigure}[b]{\columnwidth}
    \centering
    \includegraphics[width=\columnwidth]{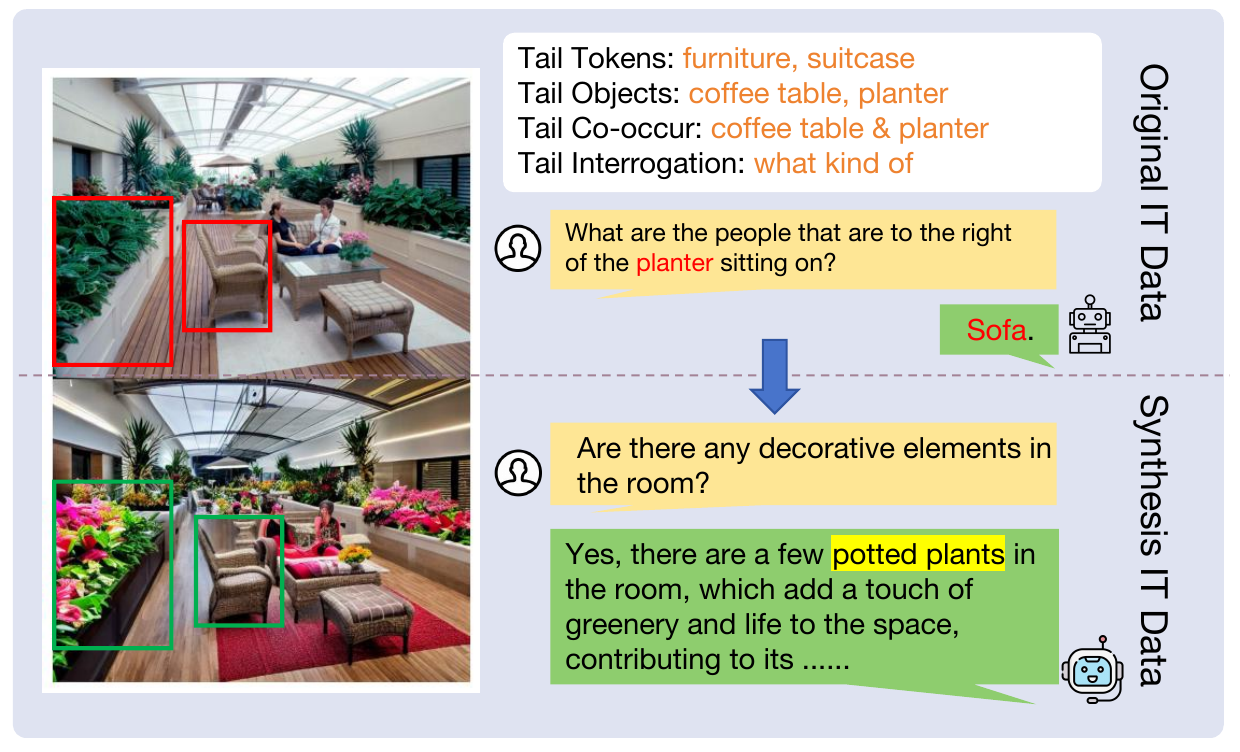}
    \caption{A furniture arrangement complemented by a variety of planters.}
\end{subfigure}

\caption{
Comparison between the original instruction-tuning (IT) data and our synthesized IT data. Tail concepts in the original data are highlighted using \red{red} boxes and fonts, whereas synthesized tail concepts are marked with \textcolor{green}{green} boxes and \colorbox{yellow}{yellow} fonts.
}
\label{fig:appendix_systhesis_qualitiveres} 

\end{figure}

We present the full qualitative results in Figure \ref{fig:appendix_qualitiveres}. LLaVA 1.5 often fails to provide accurate responses when addressing tail questions. However, with the integration of our ADR framework, the model demonstrates significant improvement in recognizing and handling tail concepts. Additionally, we showcase more examples of our synthesized data in Figure \ref{fig:appendix_systhesis_qualitiveres}. This synthesis process enriches the tail data with additional instances, effectively boosting the model’s generalization and performance in underrepresented scenarios.

\begin{table*}[p]

\caption{\textbf{Comparison of models trained with different approaches across multiple benchmarks.} IT represents the number of training instances used during instruction tuning. \textcolor{blue}{+DR} signifies performance after the data rebalancing stage, and \textcolor{red}{+DS} indicates performance after the data synthesis stage, with the number following DS denoting the augmentation volume from the DS stage. Benchmark names are abbreviated due to space constraints. The best results are indicated in \textbf{bold}.}

\label{tab:detailed_main_results}
\centering
\resizebox{\textwidth}{!}{
\begin{tabular}{lc|ccccccccccc }
\toprule
Method & IT* & VQA$^\text{OK}$ & SEED$^{\text{2}}$ & QB$^\text{2}$ & MMS & MME$^\text{P}$ & SQA$^\text{I}$ & MMMU & VQA$^\text{T}$ & GQA & MMB & VQA$^\text{v2}$ \\
\midrule
LLaVA 1.5 & 665.0K & 53.2 & 48.7 & 47.3 & 33.5 & 1510.7 & 69.3 & 35.3 & 46.0 & 61.9 & 64.3 & 76.6\\ 
\hspace{0.3cm}\textcolor{blue}{+DR} & 581.0K & 55.3 & 57.2 & 46.8 & 33.8 & 1470.6 & 69.5 & 34.8 & 46.0 & 62.8 & 65.5 & 76.9 \\ 
\hspace{0.3cm}\textcolor{blue}{+DR} \textcolor{red}{+DS} & 665.0K & \textbf{57.4} & \textbf{57.4} & \textbf{49.6} & \textbf{35.5} & \textbf{1512.8} & \textbf{70.4} & 36.7 & \textbf{47.2} & \textbf{62.9} & 65.0 & 76.9 \\ 
\hline
\hspace{0.3cm}\textcolor{red}{+DS} 25K & 690.0K & 56.2 & 47.5& 47.9& 34.5& 1486.0& 68.7& 36.0& 47.1& 62.8& 66.3 & \textbf{77.2} \\ 
\hspace{0.3cm}\textcolor{red}{+DS} 50K & 715.0K & 57.3 & 47.3& 47.7& 35.2& 1472.5& 69.9& \textbf{36.9}& 47.0 & 62.7& \textbf{66.3}& 77.1 \\ 
\hspace{0.3cm}\textcolor{red}{+DS} 100K & 765.0K & 54.5& 47.2& 46.1& 34.6& 1502.7& 69.7& 36.8& 46.1& 62.5& 64.5& 76.6\\ 

\bottomrule
\end{tabular}
}
\end{table*}

\begin{table*}[p]

\caption{Full results of ablation study on different combinations of perspectives. T, O, C, and W refer to Token, Object, Co-occurrence, and Interrogation respectively. The best results are indicated in \textbf{bold}, and the second-best results are \underline{underlined}.}
\label{tab:ablation_different_combination}
\centering
\setlength{\tabcolsep}{4pt} 
\resizebox{\textwidth}{!}{  
\begin{tabular}{cccc| c | cccccccccccc }
\toprule
T & O & C & W & IT & VQA$^\text{v2}$ & VQA$^\text{T}$ & VQA$^\text{OK}$ & GQA & SQA & SQA$^\text{I}$ & REF & REF+ & FLIK & POPE & SEED & Avg. \\
\midrule

\multicolumn{4}{c|}{baseline} & 665.0K & 76.6& 46.0& 53.2& 61.9& 70.4& 69.3& 29.4& 28.5& 74.9& 86.9& 60.6 & 59.8\\ 
\checkmark & & &  & 488.1K & 76.5& 46.6& 55.3& 62.3& 70.8& 69.2& 28.5& 28.1& 73.8& 86.7& 60.2 & 59.8\\ 
 & \checkmark& &  & 197.9K & 74.6& 44.0& 50.4& 61.3& 69.9& 67.9& 30.8& 29.7& 74.1& 86.3& 59.3 & 59.0\\ 
 & &\checkmark &  & 242.4K & 75.2& 43.3& 47.3& 61.3& 70.0& 68.5& \underline{31.4}& 29.8& 76.2& 86.8& 59.0 & 59.0\\ 
 & & &\checkmark  & 176.3K & 73.9& 43.0& 46.3& 60.7& 69.5& 66.7& \textbf{32.3}& \textbf{31.7}& 71.9& 85.6& 57.4 & 58.1\\ 
\checkmark & \checkmark & &  & 534.2K & 76.7& \underline{47.1}& \underline{55.6}& 62.8& 71.4& 68.1& 30.3& 29.1& 75.4& 86.9& 60.9 & 60.4\\ 
\checkmark &  & \checkmark &  & 553.4K & 75.7& 44.5& 52.8& 62.0& 70.8& 68.4& 30.4& 29.2& 75.1& 86.4& 59.9 & 59.6\\ 
\checkmark &  & & \checkmark  & 521.5K & 75.7& 44.5& 52.8& 62.0& 70.8& 68.4& 30.4& 29.2& 75.1& 86.4& 59.9 & 59.6\\ 
 & \checkmark & \checkmark &  & 276.9K & 75.4& 44.6& 46.8& 61.7& 69.0& 66.4& 30.6& 29.4& 74.2& 87.1& 59.3 & 58.6\\ 
 &  \checkmark & & \checkmark  & 318.3K & 75.7& 44.6& 50.9& 61.8& 71.5& 69.0& 29.9& 29.0& 74.9& 86.8& 59.6 & 59.4\\ 
 & & \checkmark &  \checkmark  & 349.9K & 76.8& 46.8& 54.4& 62.5& 71.5& 68.8& 29.9& 29.2& 75.7& 86.8& \textbf{61.5} & 60.4\\ 
 & \checkmark & \checkmark &  \checkmark  & 375.9K & 76.2& 45.3& 54.4& \underline{62.8}& 70.7& 67.6& 29.7& 28.8& 74.3& 86.8& 60.1 & 59.7\\ 
 \checkmark &  & \checkmark &  \checkmark  & 575.5K & 76.8& 46.7& \textbf{56.7}& 62.4& 71.2& 68.8& 30.1& 29.1& 75.9& 87.2& \underline{61.2} & 60.6\\ 
 \checkmark &\checkmark  &  &  \checkmark  & 559.3K & 76.7& 46.9& 52.5& 62.3& \underline{71.6}& 69.2& 30.8& \underline{30.0}& \textbf{76.6}& \textbf{87.4}& 61.0 & 60.5\\ 
 \checkmark &\checkmark  & \checkmark &    & 561.5K & \underline{76.8}& \textbf{47.2}& 50.0& 62.3& \textbf{71.7}& \textbf{69.9}& 28.8& 28.1& 75.6& 86.6& 60.6 & 59.8\\ 
 \checkmark &\checkmark  & \checkmark &  \checkmark  & 581.7K & \textbf{76.9}& 46.0& 55.3& \textbf{62.8}& 71.4& \underline{69.5}& 30.2& 29.7& \underline{76.2}& \underline{87.2}& 61.0 & 60.6\\

\bottomrule
\end{tabular}
}
\end{table*}

\begin{table*}[p]
\caption{Full results of ablation study on different augmentation methods. Methods are introduced in Sec. 6.2. The best results are indicated in \textbf{bold}, and the second-best results are \underline{underlined}.}
\label{tab:ablation_different_augment}
\centering
\setlength{\tabcolsep}{4pt} 
\resizebox{\textwidth}{!}{  
\begin{tabular}{c c | cccccccccccc }
\toprule
Method & IT & VQA$^\text{v2}$ & VQA$^\text{T}$ & VQA$^\text{OK}$ & GQA  & SQA & SQA$^\text{I}$ & REF & REF+ & FLIK & POPE & SEED & Avg. \\
\midrule
ALL & 665.0K & 76.9& \textbf{47.2} & \textbf{57.4} & \underline{62.9}& \textbf{72.0}& \textbf{70.4} & 30.5& 29.9& 76.2& 86.9& \underline{61.3} & 61.1\\ 
Image Only & 665.0K & \underline{76.9}& 46.5& \underline{57.2}& 62.5& 68.8& 68.4& 30.6& 30.2& 75.9& 87.3& 53.8 & 59.8\\ 
Token Rewrite & 665.0K & \textbf{76.9}& 46.1& 49.2& 62.4& 70.6& 68.6& \textbf{32.3}& \textbf{31.3}& 0.6& 87.4& 54.1 & 52.7\\ 
TW Rewrite & 665.0K & 76.9& \underline{46.9}& 54.9& 62.5& 68.9& 68.7& 31.0& 30.3& \textbf{77.5}& \underline{87.5}& 53.7 & 59.9\\ 
PlainAug SimpAdd & 665.3K & 76.8& 46.2& 56.0& \textbf{63.0}& \underline{71.7}& 69.3& 29.3& 28.5& 74.1& 86.6& \textbf{61.7} & 60.3\\ 
PlainAug NewCap & 665.3K & 76.8& 46.7 & 54.6& 62.1& 68.5& \underline{69.4}& \underline{31.1}& \underline{30.7}& \underline{77.3}& \textbf{87.7}& 54.1 & 59.9\\

\bottomrule
\end{tabular}
}
\end{table*}

\section{Details of Analyzing Stage}
\subsection{Examples of Entities}
\label{appendix:top20}
\begin{figure*}[p]  
    \centering
    \includegraphics[width=\textwidth]{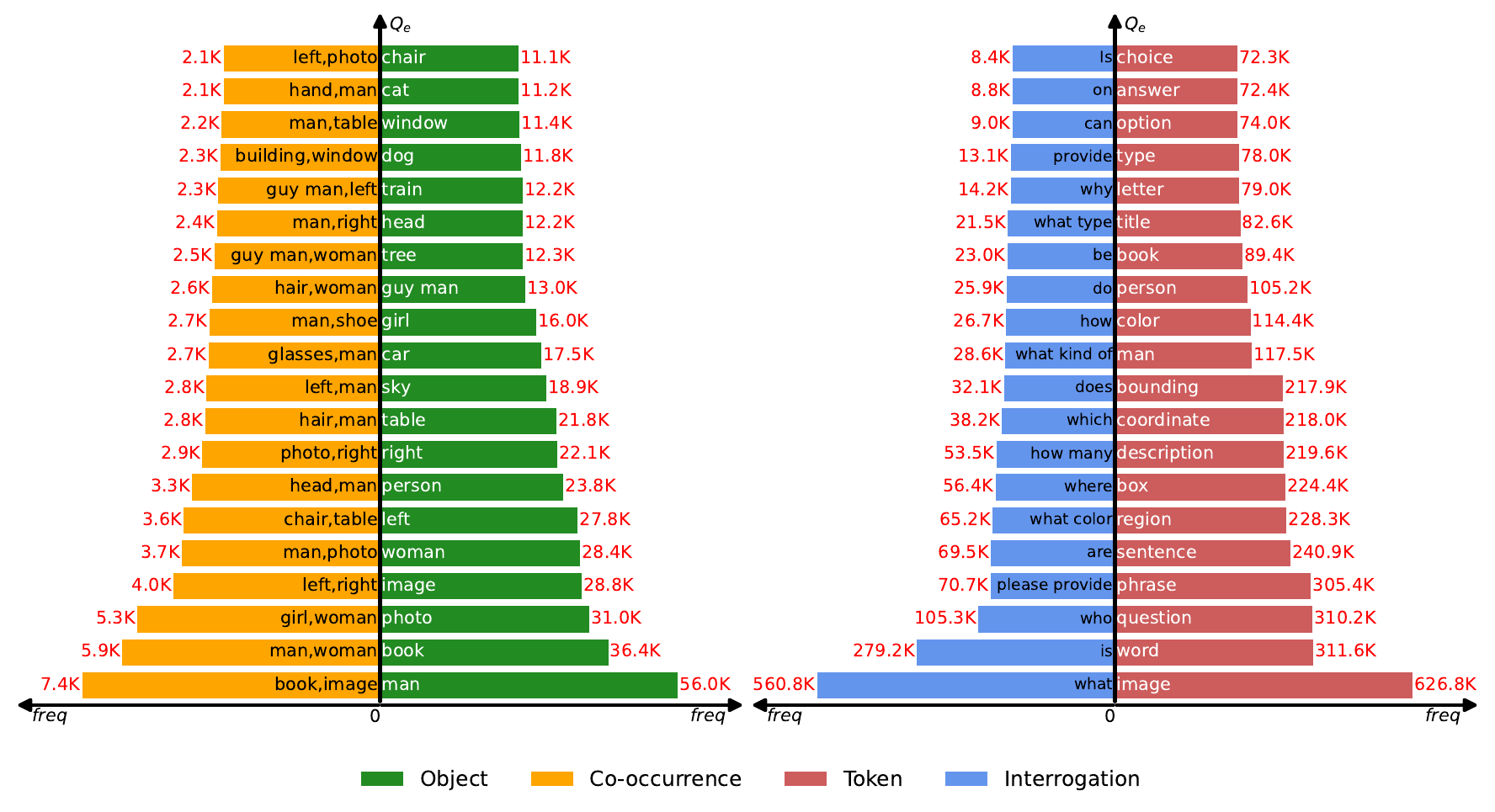}  
    \caption{Top 20 most frequent entities in the instruction-tuning dataset of LLaVA 1.5.}  
    \label{fig:top20entities}  
\end{figure*}

Different kinds of entities are extracted from four perspectives: Token, Object, Co-occurrence, and Interrogations. The top 20 frequently-shown entities from instruction-tuning data of LLaVA 1.5 are displayed in Figure \ref{fig:top20entities}.

\subsection{Implement Details of Analyzing Stage}
\label{appendix:entity_distribution_construct}

\begin{figure*}[p]
\centering    

\begin{subfigure}[b]{0.33\linewidth}
    \centering
    \includegraphics[width=\linewidth]{Styles/figures/token_mme_lt_lcs.pdf}
    \caption{MME: Tok}
\end{subfigure}
\begin{subfigure}[b]{0.33\linewidth}
    \centering
    \includegraphics[width=\linewidth]{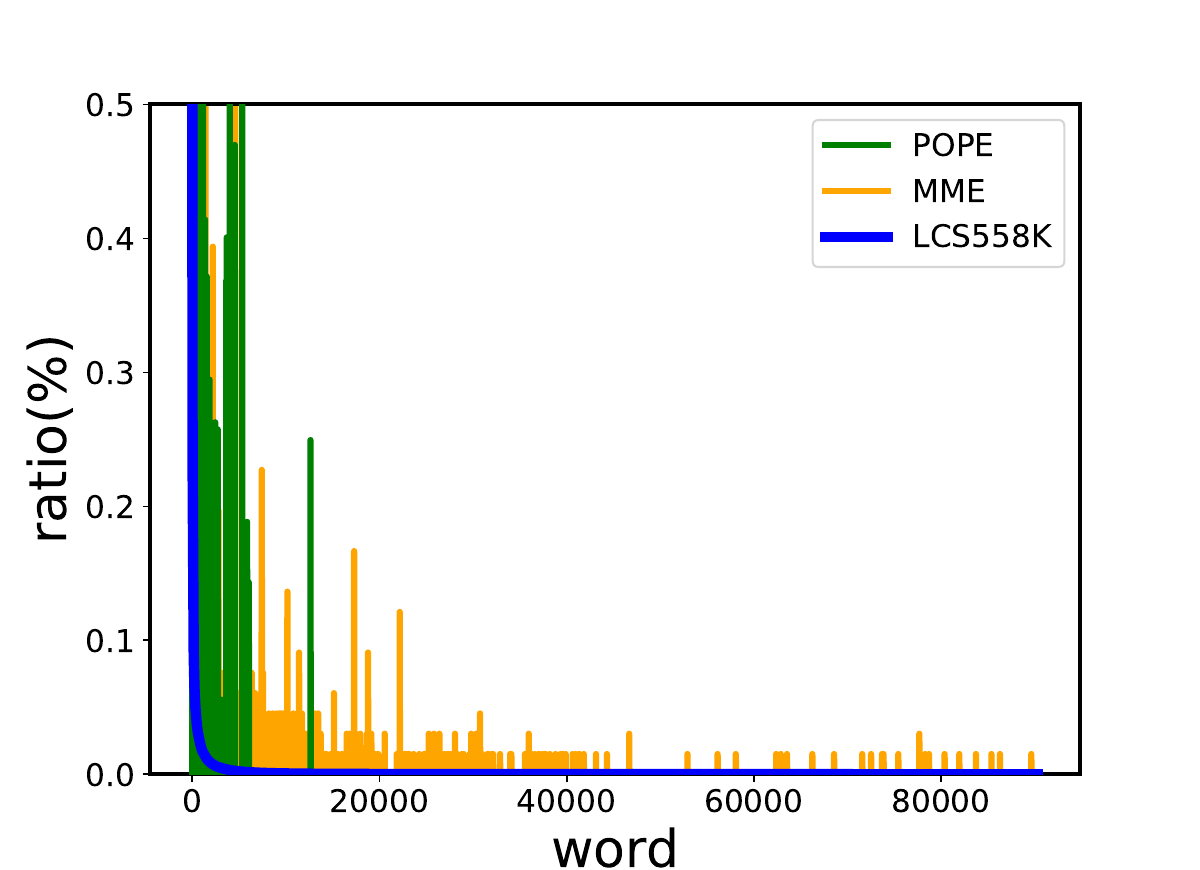}
    \caption{LCS558K: Tok}
\end{subfigure}
\begin{subfigure}[b]{0.33\linewidth}
    \centering
    \includegraphics[width=\linewidth]{Styles/figures/token_instruct_lt_pope_mme.pdf}
    \caption{InstructMix665K: Tok}
\end{subfigure}

\begin{subfigure}[b]{0.33\linewidth}
    \centering
    \includegraphics[width=\linewidth]{Styles/figures/object_mme_lt_lcs.pdf}
    \caption{MME: Obj}
\end{subfigure}
\begin{subfigure}[b]{0.33\linewidth}
    \centering
    \includegraphics[width=\linewidth]{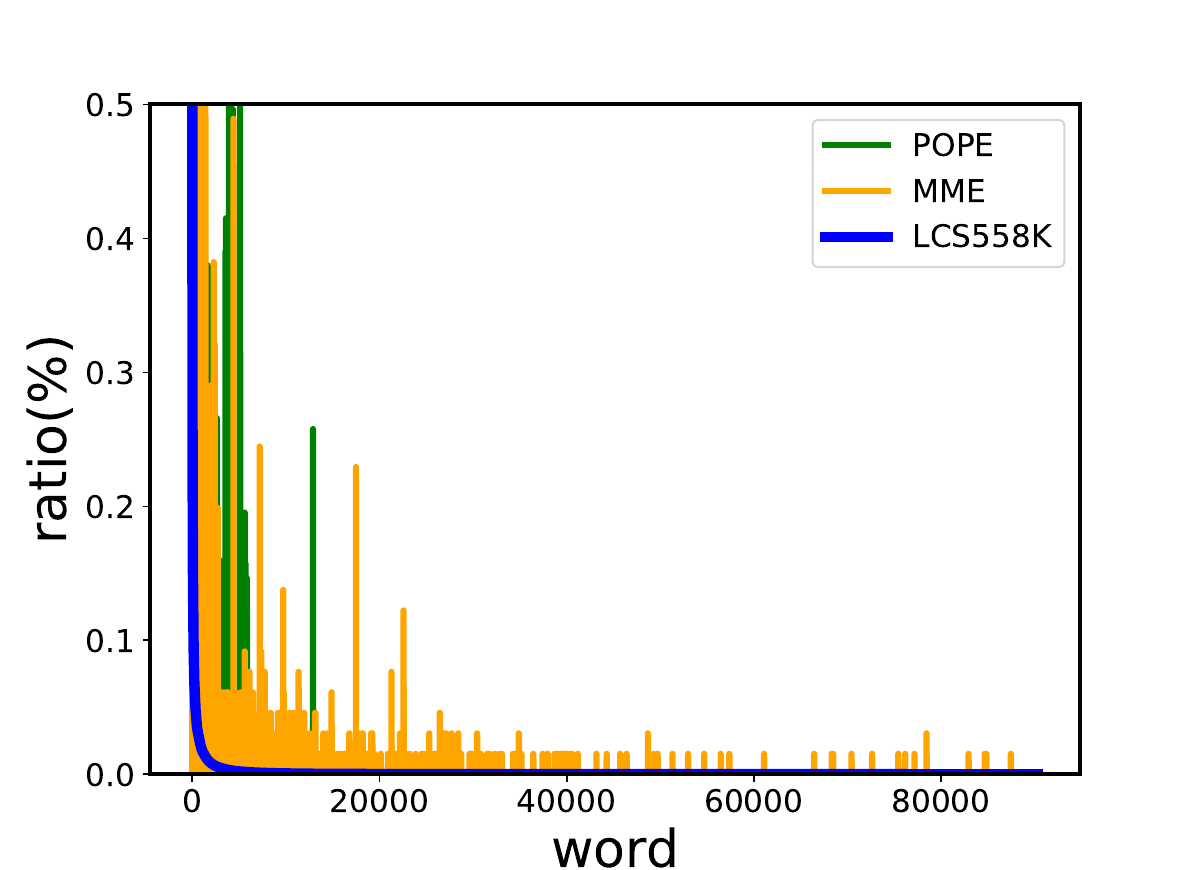}
    \caption{LCS558K: Obj}
\end{subfigure}
\begin{subfigure}[b]{0.33\linewidth}
    \centering
    \includegraphics[width=\linewidth]{Styles/figures/object_instruct_lt_pope_mme.pdf}
    \caption{InstructMix665K: Obj}
\end{subfigure}

\begin{subfigure}[b]{0.4\linewidth}
    \centering
    \includegraphics[width=\linewidth]{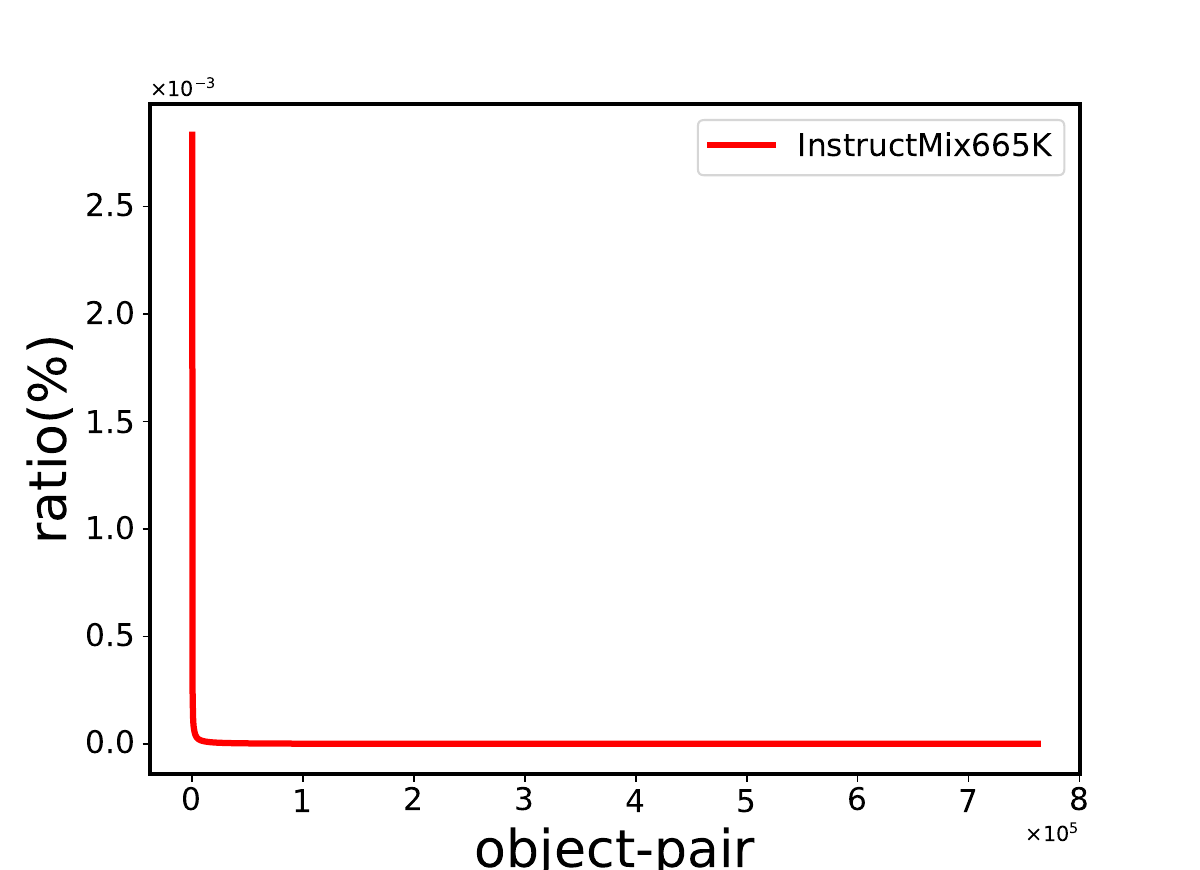}
    \caption{InstructMix665K: Co-occurrence}
\end{subfigure}
\begin{subfigure}[b]{0.4\linewidth}
    \centering
    \includegraphics[width=\linewidth]{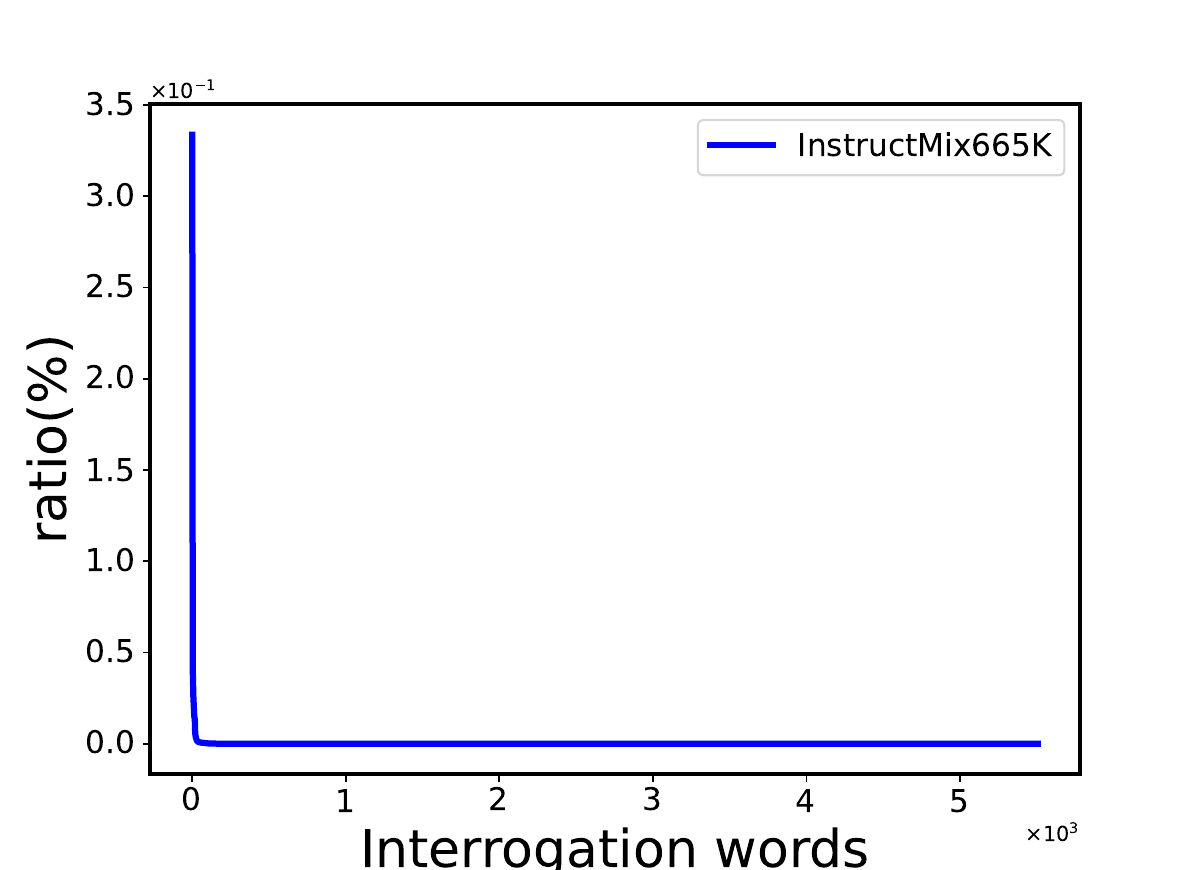}
    \caption{InstructMix665K: Interrogation}
\end{subfigure}

\caption{
Long-tail distribution in instruction-tuning and benchmark datasets. Some plots feature multiple curves, with the x-axis standardized according to the dataset mentioned in the title. Distributions from various datasets are overlaid on the same graph to emphasize the differences between them. (a) Token-level word distribution in MME \citep{fu2023MME}.
(b) Token-level word distribution in LCS558K \citep{liu2024visual}.
(c) Token-level word distribution in InstructMix665K \citep{liu2024visual}.
(d) Object-level word distribution in MME \citep{fu2023MME}.
(e) Object-level word distribution in LCS558K \citep{liu2024visual}.
(f) Object-level word distribution in InstructMix665K \citep{liu2024visual}.
(g) Co-occurrence distribution in InstructMix665K \citep{liu2024visual}.
(h) Interrogation distribution in InstructMix665K \citep{liu2024visual}.
}
\label{fig:full_long_tail_distribution_analysis}

\end{figure*}

In this work, we construct the entity distribution using both the pretraining and instruction-tuning datasets from LLaVA 1.5, specifically LCS558K and Instructmix665K. To compare the differences between training and test data further, we also incorporate portions of the distributions from POPE and MME within the same figure. The complete results are presented in Figure \ref{fig:full_long_tail_distribution_analysis}. As illustrated, all pretraining, instruction-tuning, and evaluation datasets exhibit LT issues. However, the frequency distributions of training and evaluation data differ significantly.

In the Analyzing stage, token entities are extracted using Stanza\footnote{stanza: \href{https://stanfordnlp.github.io/stanza/}{link}} \cite{qi2020stanza} as the POS parser. For object entities, we initially use LLaMA 3 70B Instruct\footnote{\label{footnote:llama3}meta-llama/Meta-Llama-3-70B-Instruct: \href{https://huggingface.co/meta-llama/Meta-Llama-3-70B-Instruct}{link}} \cite{dubey2024llama3} to detect potential object-related vocabulary, followed by GroundingDINO\footnote{IDEA-Research/grounding-dino-base: \href{https://huggingface.co/IDEA-Research/grounding-dino-base}{link}} \cite{liu2023groundingdino} to extract actual objects from the image. For co-occurrence distribution construction, we use Neo4j\footnote{Enterprise version 5.19.0: \href{https://neo4j.com/release-notes/database/neo4j-5/}{link}} to create an undirected graph. To construct interrogation entity distributions, we utilize LLaMA 3 70B Instruct\footref{footnote:llama3} \cite{dubey2024llama3} to extract interrogation words.

\subsection{Analysis of Failed Cases}
\label{appendix:failed_cases}

We experiment to observe the distribution location of failed cases. We first extract all entities within the failed cases and calculate the max, min, and average location of these entities in the pertaining distribution. Also, we calculate the distribution locations of the correct cases as well to compare. The results are shown in Table \ref{tab:wrong_right_locations}. As shown in the table, it is easy to discover that the failed cases are positioned further behind the correct ones in the distribution.

\begin{table*}[p]
  \caption{Distribution locations of entities in correct and incorrect answers for POPE and MME, generated by LLaVA 1.5. “Tok,” “Obj,” and “Co” refer to Token, Object, and Co-occurrence, respectively, while “W” and “C” represent wrong and correct answers, respectively. The gray rows (\textcolor{gray}{\rule{5pt}{5pt}}) indicate the relative displacement of incorrect concepts in the distribution compared to correct concepts.}
  \label{tab:wrong_right_locations}
  \centering
  \resizebox{\textwidth}{!}{
  \begin{tabular}{l llllll llllll}
    \toprule
    \multirow{2}{*}[-0.66ex]{\textbf{Methods}} & \multicolumn{6}{c}{\textbf{MME}} & \multicolumn{6}{c}{\textbf{POPE}} \\
    \cmidrule(lr){2-7}\cmidrule(lr){8-13}
     & \textbf{Tok-C} & \textbf{Tok-W}& \textbf{Obj-C} & \textbf{Obj-W} & \textbf{Co-C} & \textbf{Co-W}  & \textbf{Tok-C} & \textbf{Tok-W}& \textbf{Obj-C} & \textbf{Obj-W} & \textbf{Co-C} & \textbf{Co-W} \\ 
    \midrule
    Max & 9738 & 10377 & 2708 & 3222 & 247315 & 257107 & 2242 & 2772 & 1085 & 1100 & 130043 & 141722 \\
    
   \rowcolor{gray!30}  & &\textcolor{red}{+639} && \textcolor{red}{+514} && \textcolor{red}{+9792} && \textcolor{red}{+30} && \textcolor{red}{+15} && \textcolor{red}{+11679} \\
   
    Min & 1 & 1 & 60 & 131 & 12732 & 20741 & 1 & 1 & 17 & 21 & 926 & 1033 \\
    
   \rowcolor{gray!30} & & \textcolor{green}{+0} && \textcolor{red}{+71} && \textcolor{red}{+8009} && \textcolor{green}{+0} && \textcolor{red}{+4} && \textcolor{red}{+107} \\
   
    Mean & 1035 & 1068 & 842 & 1035 & 71123 & 79104 & 313 & 340 & 319 & 336 & 27457 & 30989 \\
    
   \rowcolor{gray!30} & & \textcolor{red}{+33} && \textcolor{red}{+193} && \textcolor{red}{+7981} && \textcolor{red}{+27} && \textcolor{red}{+17} && \textcolor{red}{+3532} \\
   
    \bottomrule
  \end{tabular}
  }
\end{table*}

\section{Details of our ADR Approach}
\label{appendix:method}

\subsection{Data Rebalancing Method}
The algorithm for our data rebalancing method is detailed in Algorithm \ref{algo:head_distribution_balance}. Initially, we calculate the sampling probability for each entity using the reverse distribution $Q^r$ and a threshold $\tau$. Entities with higher frequencies are assigned lower sampling probabilities, reducing the likelihood of overrepresented entities being selected. We then iterate over the entire dataset, leveraging these probabilities to filter out overrepresented instances. For each data instance $d$, we assess all four perspectives via random sampling. If an entity within a perspective is sampled, the perspective is marked as ``pass''. Instances with a number of passed perspectives greater than $n_p$ are retained; otherwise, they are discarded.

\subsection{Implement Details of Data Synthesis Stage}
During the Data Synthesis (DS) stage, we use ControlNet\footnote{lllyasviel/ControlNet: \href{https://huggingface.co/lllyasviel/ControlNet}{link}} \cite{zhang2023controlnet} to generate images that closely resemble those containing tail concepts. To produce high-quality captions for the generated images, we employ ShareCaptioner\footnote{Lin-Chen/ShareCaptioner: \href{https://huggingface.co/Lin-Chen/ShareCaptioner}{link}} \cite{chen2023sharegpt4v}. Finally, we leverage LLaMA 3 70B Instruct \cite{dubey2024llama3} to expand the captions into detailed conversations.

\begin{algorithm}[!t]

    \caption{Pseudo Code for \textbf{D}ata \textbf{R}esampling}
    \label{algo:head_distribution_balance}
    \renewcommand{\algorithmicrequire}{\textbf{Input:}}
    \renewcommand{\algorithmicensure}{\textbf{Output:}}

\begin{lstlisting}[xleftmargin=2em,]
# D: raw training set;
# C: target perspectives list
# tau: the threshold for entities; 
# D_bal: the rebalanced data, a.k.a. D*;
# n_p, alpha: hyperparameters
D_bal=[]           
for pers in C:     # build prob dict
    entity_dist = entity_distribution_construction(D,pers)
    prob_dict[pers] = {ent:tau[pers]/entry_dist[ent] for ent in entry_dict.keys()}
for instance in D: #  data rebalancing
    pass_cnt = 0   
    for pers in C:
        for entity in instance['entity'][pers]:
            if random.random() < prob_dict[pers][entity]:
                pass_cnt += 1
                break
    if pass_cnt > n_p and random.random() < alpha:
        D_bal.append(instance)
\end{lstlisting}
\end{algorithm}

\section{Prompts}

\subsection{Object Information Extraction}
\label{appendix:prompts_obje}
\begin{figure*}[p]  
    \centering
    \includegraphics[width=\textwidth]{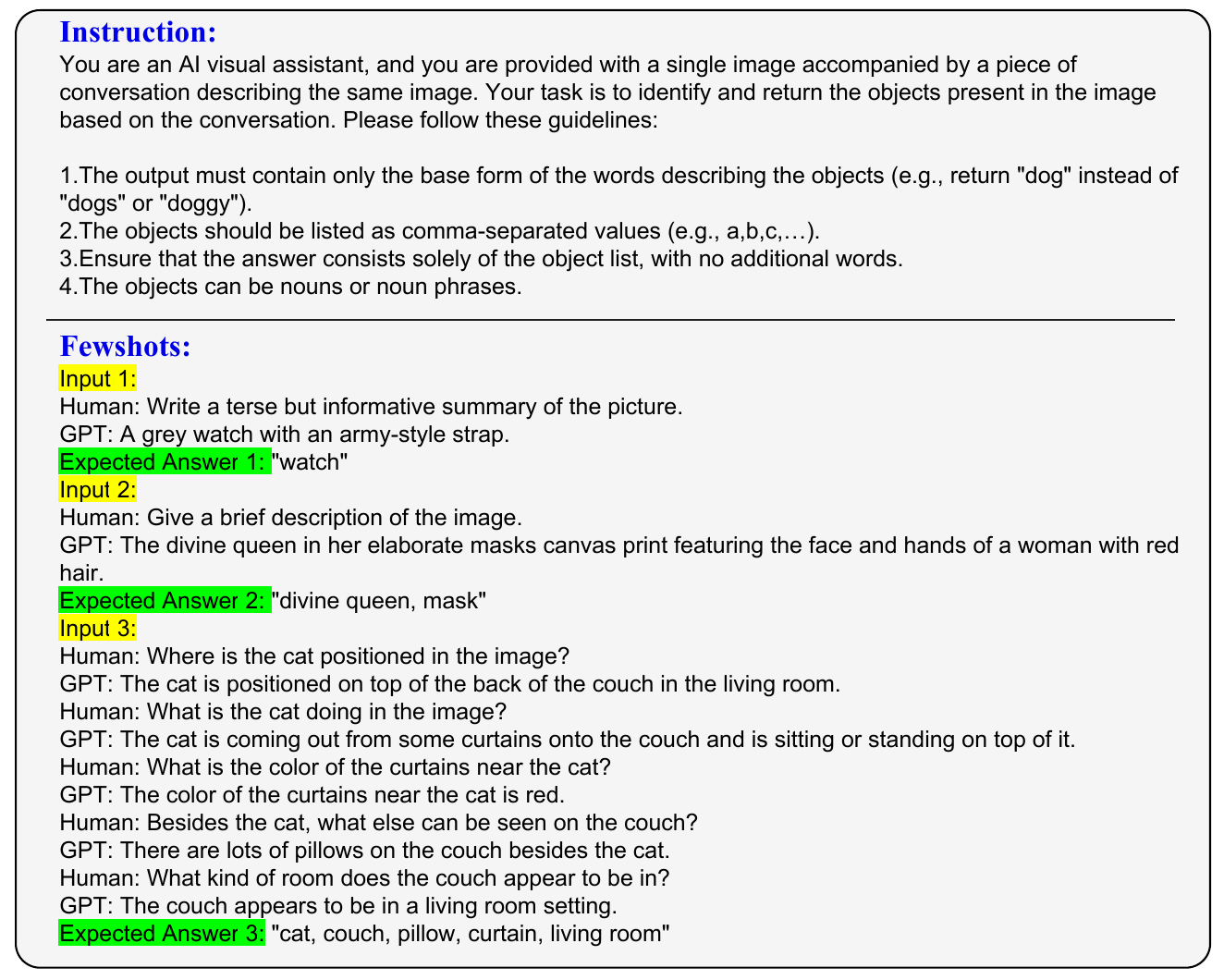}  
    \caption{Complete prompts used to guide the language model in extracting object information.} 
    \label{fig:prompt_extract_obj}  
\end{figure*}

In this section, we release all of our prompts for guiding LLMs to do specific tasks. Firstly during the analyzing stage, we utilize the LLMs to extract object information from the text within data instances at the very first step during object entity extraction. This part of the prompt we used to guide LLMs is illustrated in Figure \ref{fig:prompt_extract_obj}.

\subsection{Conversation Rewrite}
\label{appendix:prompts_rewrite}
\begin{figure*}[p]  
    \centering
    \includegraphics[width=\textwidth]{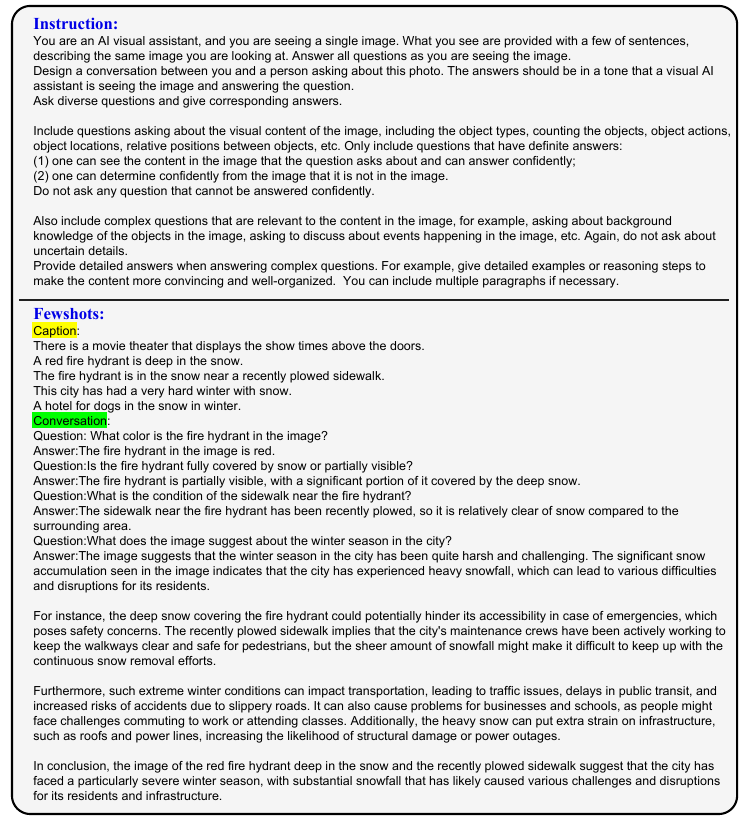}  
    \caption{Complete prompts used to guide the language model in converting captions into conversation instructions.} 
    \label{fig:prompt_cap2conv}  
\end{figure*}

We leverage LLaMA3 70B Instruct \citep{dubey2024llama3} to rewrite our conversations. During the Data Synthesis (DS) Stage, synthetic data and captions are generated using diffusion models and captioning models. Once the image and its corresponding caption are obtained, we employ the LM to transform the caption into a conversation. The prompt used to guide the LM is shown in Figure \ref{fig:prompt_cap2conv}.

\begin{figure*}[p]  
    \centering
    \includegraphics[width=\textwidth]{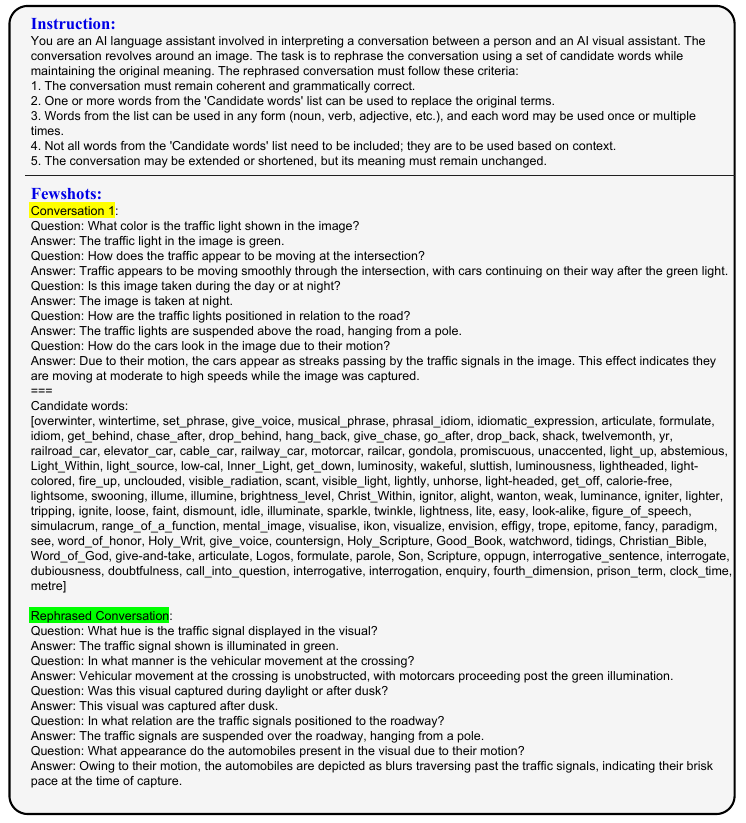}  
    \caption{Complete prompts used to guide the language model in rewrite conversation instructions using given tokens.} 
    \label{fig:prompt_token_rewrite}  
\end{figure*}

Moreover, during the language data synthesis process in the DS stage, we also utilize LLMs to rewrite conversations using the provided tail tokens. The corresponding prompts are shown in Figure \ref{fig:prompt_token_rewrite}. Additionally, we rewrite conversations containing tail tokens or interrogation entities (TWR in the ablation study or Section 6.2). As this task closely resembles standard rephrasing tasks with similar prompts, we will not elaborate on it further here.

\clearpage

